\documentclass{article}

\usepackage{microtype}
\usepackage{graphicx}
\usepackage{subfigure}
\usepackage{booktabs} 

\usepackage{xcolor}
\definecolor{y}{RGB}{255, 250, 205}
\definecolor{p}{RGB}{245, 234, 240}
\definecolor{b}{RGB}{224, 255, 255}
\definecolor{g}{RGB}{193, 255, 193}

\usepackage{algorithm}
\usepackage{makecell}

\definecolor{PineGreen}{HTML}{008B00}
\definecolor{BrickRed}{HTML}{B22222}

\newcommand{\SOTAmethod}[0]{{LaFTer}~}

\newcommand{\ourmethodCC}[0]{{CPL}~}

\usepackage{hyperref}

\usepackage[accepted]{icml2024}
\usepackage{amsmath}
\usepackage{bbm}
\usepackage{amssymb}
\usepackage{mathtools}
\usepackage{amsthm}
\usepackage{subcaption}
\usepackage{graphicx}
\usepackage{caption}
\usepackage{multirow}
\usepackage{pifont} 
\usepackage[capitalize,noabbrev]{cleveref}

\theoremstyle{plain}

\theoremstyle{definition}

\theoremstyle{remark}

\newcommand{\bs}{\boldsymbol}
\usepackage{enumitem}
\usepackage{colortbl}
\definecolor{bgreen}{RGB}{0,170,0}
\definecolor{bred}{RGB}{220,0,0}
\definecolor{mydarkblue}{RGB}{0,0,150}
\definecolor{Gray}{gray}{0.93}
\hypersetup{
    colorlinks=true,
    linkcolor=bred,
    citecolor=mydarkblue,
    filecolor=bred,
    urlcolor=mydarkblue
}

\icmltitlerunning{Candidate Pseudolabel Learning: Enhancing Vision-Language Models by Prompt Tuning with Unlabeled Data}

\begin{document}

\twocolumn[
\icmltitle{Candidate Pseudolabel Learning: Enhancing Vision-Language Models by Prompt Tuning with Unlabeled Data}



\icmlsetsymbol{equal}{*}

\begin{icmlauthorlist}
\icmlauthor{Jiahan Zhang}{equal,Zhang}
\icmlauthor{Qi Wei}{equal,wei}
\icmlauthor{Feng Liu}{Liu}
\icmlauthor{Lei Feng}{Zhang}

\end{icmlauthorlist}

\icmlaffiliation{Zhang}{Singapore University of Technology and Design, Singapore}
\icmlaffiliation{wei}{Nanyang Technological University, Singapore}
\icmlaffiliation{Liu}{University of Melbourne, Australia}


\icmlcorrespondingauthor{Lei Feng}{feng\_lei@sutd.edu.sg}

\icmlkeywords{Machine Learning, ICML, Prompt Learning, Pseudolabel, VLMs, Vision-Language}

\vskip 0.3in
]

\printAffiliationsAndNotice{\icmlEqualContribution} 

\begin{abstract}
Fine-tuning vision-language models (VLMs) with abundant unlabeled data recently has attracted increasing attention. Existing methods that resort to the pseudolabeling strategy would suffer from heavily incorrect hard pseudolabels when VLMs exhibit low zero-shot performance in downstream tasks. To alleviate this issue, we propose a \textbf{C}andidate \textbf{P}seudolabel \textbf{L}earning method, termed \textbf{CPL}, to fine-tune VLMs with suitable candidate pseudolabels of unlabeled data in downstream tasks.
The core of our method lies in the generation strategy of candidate pseudolabels, which progressively generates refined candidate pseudolabels by both intra- and inter-instance label selection, based on a confidence score matrix for all unlabeled data. This strategy can result in better performance in true label inclusion and class-balanced instance selection. In this way, we can directly apply existing loss functions to learn with generated candidate psueudolabels. Extensive experiments on nine benchmark datasets with three learning paradigms demonstrate the effectiveness of our method. Our code can be found \href{https://github.com/vanillaer/CPL-ICML2024}{here}.
\end{abstract}

\section{Introduction}
\label{sec:Introduction}
Recent studies in large pre-trained vision-language models (VLMs) \cite{radford2021learning,li2022blip,yuan2021florence} have demonstrated promising zero-shot performance. Nonetheless, previous research \cite{zhou2022conditional,zhou2022coop,zhang2024vision} indicated that substantial labeled data is still necessary to further improve the performance of VLMs for adaptation on various downstream tasks. This requirement for adaptation would cause considerable labeling costs, as labeled data is hard to obtain.

In response to this challenge, recent studies \cite{menghini2023enhancing,huang2022unsupervised,lai2023padclip,tanwisuth2023POUF,shu2022testtime} have shifted their focus towards scenarios with abundant unlabeled data, aiming to exploit the inherent zero-shot ability of VLMs. These studies can reduce the dependency on labeled data for adapting VLMs to downstream tasks.
Besides, previous studies also showed that fine-tuning VLMs with pseudolabels generated by the zero-shot ability of VLMs is an effective approach for exploiting unlabeled data \cite{huang2022unsupervised,menghini2023enhancing,mirza2023lafter}.
However, the performance of existing methods heavily relies on the accuracy of the generated hard pseudolabels. When VLMs exhibit diminished zero-shot abilities in certain downstream tasks, the performance of these methods would significantly deteriorate.

\begin{figure}[t]
\centering     
\subfigure[]{\label{fig:matrixa}\includegraphics[width=51mm]{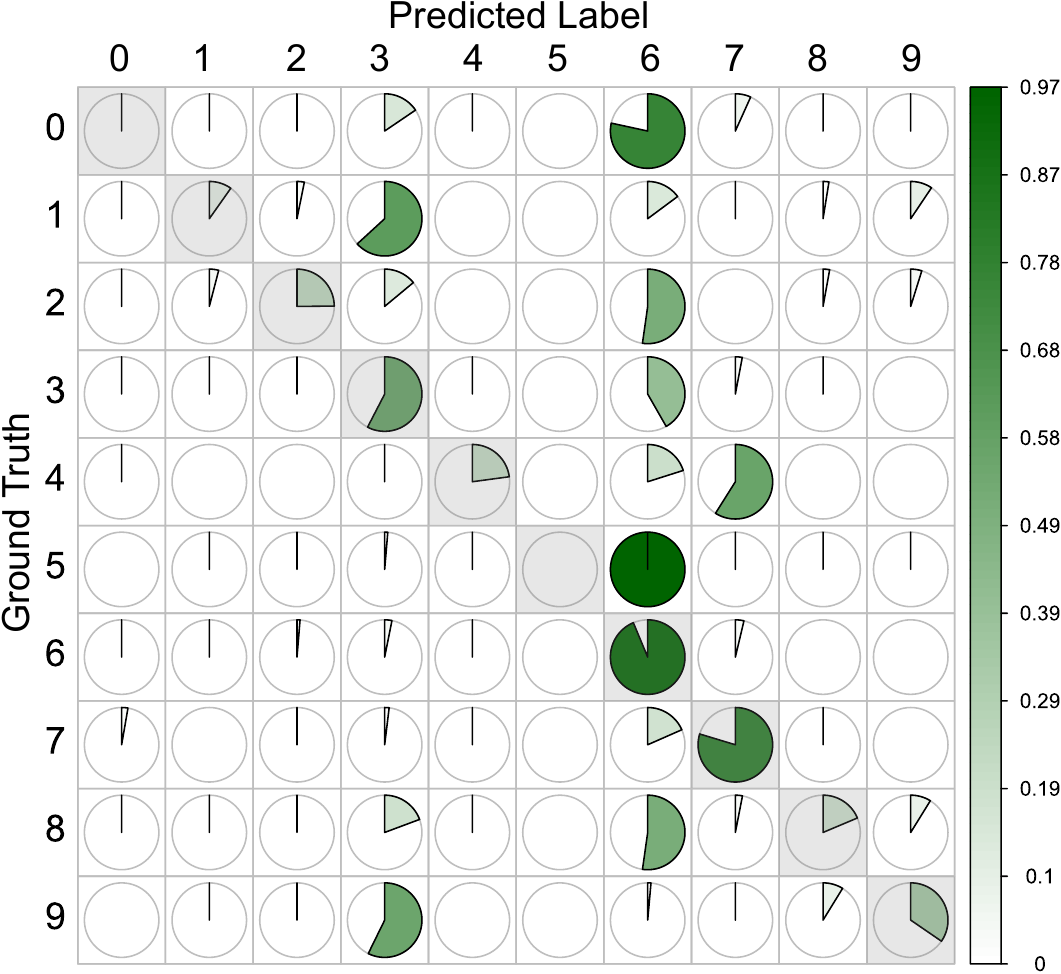}}
\hspace{0.5mm}
\subfigure[]{\label{fig:matrixb}\includegraphics[width=29mm]{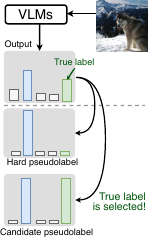}}\\
\vspace{-3mm}
\caption{(a) Confusion matrix between true labels and hard pseudolabels of dataset EuroSAT, where incorrect and imbalanced pseudolabels are always generated. 
(b) An example illustration of a set of candidate pseudolabels, which consists of classes with the top-2 highest confidence scores.
}
\vspace{-3mm}
\end{figure}

To illustrate this issue, we conducted a pilot experiment to empirically demonstrate the deficiency of hard pseudolabels. Specifically, we leveraged CLIP \cite{radford2021learning} to generate hard pseudolabels on the EuroSAT dataset \cite{helber2019eurosat} and subsequently calculated the confusion matrix between the ground-truth labels and predicted class labels. The results, as shown in \cref{fig:matrixa}, reveal that a substantial number of samples from other categories are incorrectly predicted as class 6, while fewer samples are classified into classes 0 and 5. Fine-tuning VLMs on such a training set with a significant number of incorrect and imbalanced pseudolabels inevitably leads to inferior performance.

\begin{figure}[t]
\centering     
\label{fig:curvea}\includegraphics[width=0.999\linewidth]{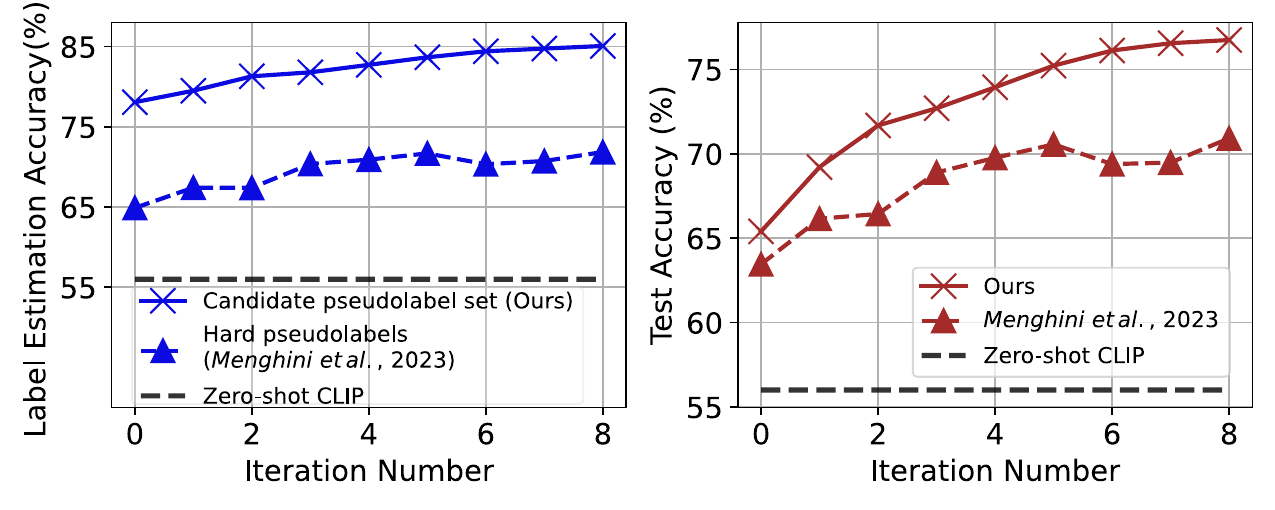}
\vspace{-3mm}
\caption{Our candidate pseudolabel learning (CPL) significantly surpasses hard pseudolabel learning \cite{menghini2023enhancing} on the RESISC45 dataset in terms of {\color{blue}label estimation accuracy} (label estimation accuracy is defined as the rate at which the true label is included in the pseudolabels), leading to improved performance on {\color{bred} test accuracy}. 
}
\vspace{-3mm}
\label{fig:prediction}
\end{figure}

Motivated by the idea of multiple annotations in crowdsourcing \cite{hossain2015crowdsourcing,li2023beyond}, we aim to provide \emph{a set of pseudolabels} that can potentially be the true label (dubbed \emph{candidate pseudolabels} in this paper), instead of merely considering a single hard pseudolabel. 
To form the set of candidate pseudolabels for each instance, we select the classes with the top-$k$ highest prediction confidences, which would probably contain the true label even though the (top-1) prediction is incorrect (see Figure \ref{fig:matrixb}). We also demonstrate the advantages of using candidate pseudolabels in \cref{fig:prediction}, where the blue lines represent the true label estimation accuracy of candidate pseudolabel learning and hard pseudolabel learning. On the other side, the red lines represent the trend of test accuracy for both methods throughout the iterations. From \cref{fig:prediction}, we can find that with training proceeds, the true label is given by the candidate pseudolabel with a probability of almost 85\%, while it is around 70\% for hard pseudolabeling. In addition, we can also find that the test accuracy of the model can be continually improved as the iteration increases.

In this paper, we propose a Candidate Pseudolabel Learning (CPL) method to fine-tune VLMs with suitable candidate pseudolabels of unlabeled data in downstream tasks.
The core of our CPL framework lies in the generation strategy of candidate pseudolabels, which needs to progressively generate refined candidate pseudolabels during the fine-tuning process.
Specifically, we construct a confidence score matrix encompassing all unlabeled data. Based on this matrix, we take into account two aspects, including intra- and inter-instance label selection. The simultaneous consideration of the two aspects can ensure that the true label is selected in the set of candidate pseudolabels to a large extent.
Meantime, it can effectively mitigate the overwhelming influence that dominant classes may exert on the generation of pseudolabels, thereby ensuring a balanced and accurate representation of classes.

Based on this novel pseudolabel structure, we transform the multiclass classification problem into the problem of learning with multiple candidate labels \cite{luo2010learning,cour2011learning}. In this way, we can fine-tune VLMs with candidate pseudolabels using popular loss functions for learning with multiple candidate labels \cite{feng2020provably,wen2021leveraged,zhang2021exploiting}. In our CPL method, the fine-tuning of the model and the update of candidate pseudolabels are conducted iteratively, mutually benefiting each other. 
We conduct extensive experiments across three learning paradigms (unsupervised learning, semi-supervised learning, and transductive zero-shot learning) and two prompt-tuning paradigms (textual and visual). Experimental results demonstrate that our proposed method consistently achieves state-of-the-art performance.

\section{Related Work}
\label{sec:Related Works} 
\subsection{Vison-Language Models}
Recently, Vision Language Models, such as CLIP \cite{radford2021learning}, ALBEF \cite{li2021align}, BLIP \cite{li2022blip}, and Flamingo \cite{alayrac2022flamingo}, pre-trained on large-scale image-text data, have achieved significant success \cite{zhang2024vision}. These models are capable of zero-shot image classification. Besides, the performance of VLMs can be further enhanced by fine-tuning with annotated data from downstream datasets. For instance, CoOp \cite{zhou2022coop} learns prompt vectors by minimizing prediction errors using the cross-entropy loss. Tip-Adapter \cite{zhang2021tip} employs additional adapter modules for parameter-efficient fine-tuning on downstream datasets. In this paper, we primarily focus on the performance of CLIP, a representative VLM, in downstream tasks.

\subsection{Prompt Tuning}
Prompt tuning is a technique that can enhance the performance of large pre-trained models in specific downstream tasks through efficient parameter fine-tuning. The common types of prompt tuning include text-based \cite{zhou2022coop,zhou2022conditional,ge2023domain} and visual prompt tuning \cite{bahng2022exploring,jia2022visual}. Text-based prompt tuning \cite{zhou2022coop} employs continuous optimization strategies to optimize a set of continuous vectors, thereby eliminating the need for manually designed discrete prompt texts. visual prompt tuning \cite{jia2022visual} offers an efficient alternative to complete fine-tuning of transformer models by introducing a minimal number of trainable parameters in the visual input. In classification tasks, prompt tuning necessitates training on a small number of labeled examples for each class. 
In this paper, we mainly explore candidate pseudolabels in visual prompt tuning \cite{jia2022visual} and text prompting tuning \cite{zhou2022coop} to enhance the performance of VLMs when unlabeled data is available.

\subsection{Learning from Unlabeled Data}
In real-world applications, we often have access to a substantial amount of unlabeled data for downstream tasks. This motivates us to devise effective methods for utilizing such data. In semi-supervised learning \cite{sohn2020fixmatch,xu2021dash,Wei2024,zhang2021flexmatch}, pseudolabeling is a widely studied and adopted technique. Grounded in the principle of entropy minimization \cite{grandvalet2004semi}, it typically selects the most reliable samples from unlabeled data based on the confidence for inclusion in training. However, this pseudolabeling strategy has been found challenging to apply directly to the zero-shot predictions of VLMs as it struggles to effectively estimate the most accurate samples from unlabeled data \cite{huang2022unsupervised}. In previous research on pseudolabeling for VLMs, \citet{huang2022unsupervised} initially proposed generating more reliable offline pseudolabels by selecting multiple examples with the highest confidence for each category. On the other hand, \citet{menghini2023enhancing} proposed updating pseudolabels iteratively while still selecting the most reliable samples for each category. In this paper, we propose a novel candidate pseudolabel generation strategy that aims to improve the label estimation accuracy of pseudolabels, thereby enhancing the performance of VLMs when adapting to downstream tasks with unlabeled data.

\section{Methodology}
\label{sec:Methodology}

\textbf{Problem Definition.} 
In this paper, our objective is to fine-tune VLMs using downstream unlabeled data drawn from a $d$-dimensional feature space represented as $\mathcal{X} \subseteq \mathbb{R}^d$. The corresponding label space for all the downstream data is denoted as $\mathcal{Y} = \{1,..., C\}$, indicating that we are considering a $C$-class classification problem.
Specifically, our focus is on exploiting the inherent zero-shot capability of VLMs to generate pseudolabels for unlabeled data. We consider three commonly encountered learning paradigms associated with abundant unlabeled data, including Semi-supervised Learning (SSL) \cite{sohn2020fixmatch,cascante2021curriculum}, Transductive Zero-shot Learning (TRZSL) \cite{wan2019transductive,gao2020zero}, and Unsupervised Learning (UL) \cite{noroozi2016unsupervised,schmarje2021survey}.

\textbf{Motivation.} For pseudolabel generation, previous methods \cite{huang2022unsupervised, menghini2023enhancing} rely on confidence ranking to select the most confident samples for each class. However, the utilization of hard pseudolabels may inadvertently amplify the effects of lower prediction accuracy for certain categories. In this paper, we draw inspiration from the concept of the \emph{multiple annotations} in crowdsourcing, constructing a set of potential true labels for model learning. 
Intuitively, the advantages of candidate pseudolabels over hard pseudolabels can result in \emph{more precise label estimation}, implying that the candidate pseudolabels can better encapsulate the true label within its candidate set. This is illustrated and contrasted in Figure \ref{fig:prediction}.

\textbf{Overview.}
The overall workflow of our method can be divided into three steps: 
\ding{182} The generation of candidate pseudolabels for unlabeled training data (see Sec. \ref{sec:subsec3.1}). Considering both intra- and inter-instance perspectives, we selectively construct the training set with candidate pseudolabels from the unlabeled dataset $D_{\rm UL}$, while keeping balanced quantities among varying classes. 
\ding{183} Based on the candidate pseudolabel, we transform the conventional multi-classification problem into a problem akin to learning in a set of candidate labels (see Sec. \ref{sec:subsec3.2}), which has been extensively studied in the field of partial-label learning \cite{cour2011learning}. Therefore, any loss function designed for partial-label learning can be employed to update the model's parameters.
\ding{184} We iterate the preceding steps until the model's parameters are optimized for downstream tasks. This iterative process facilitates the progressive refinement and optimization of candidate pseudolabels.

\subsection{Scheme for Generating Candidate Pseudolabels}
\label{sec:subsec3.1}
\begin{figure}[t]
    \centering 
\includegraphics[width=1\linewidth]{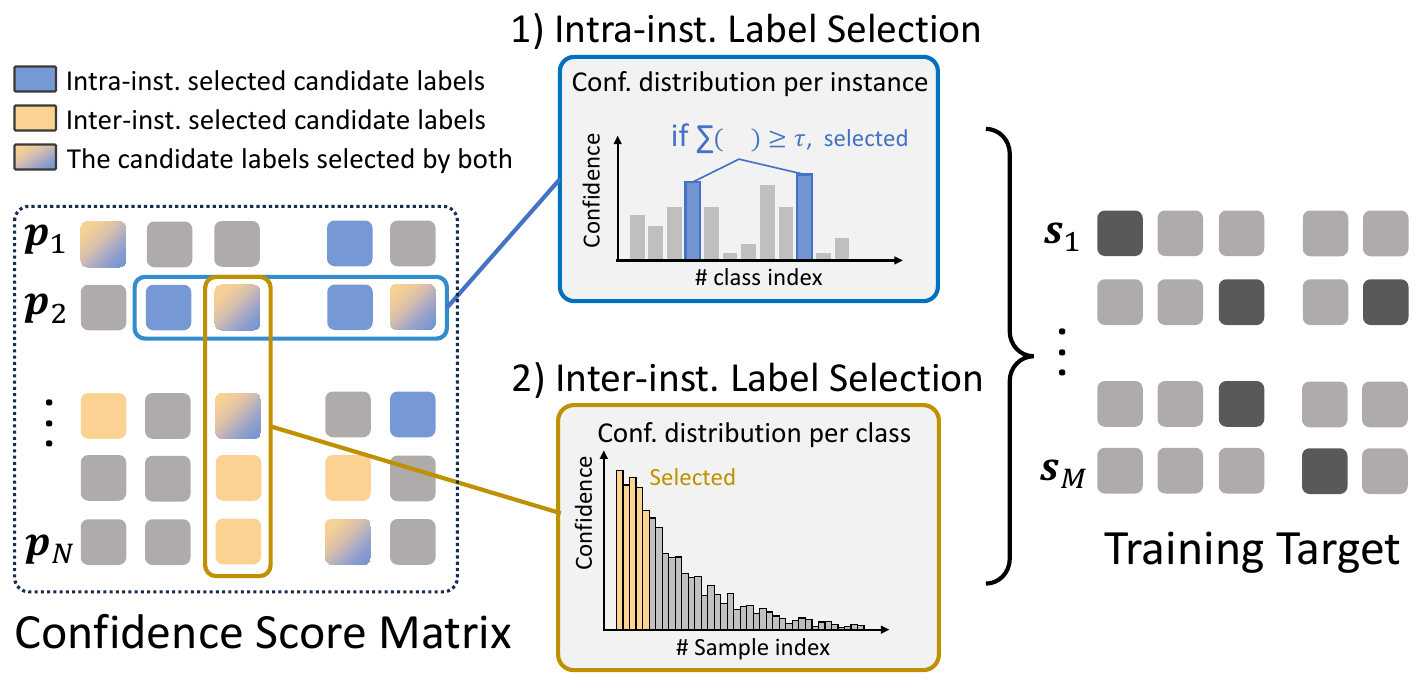}
    \caption{Illustration of the training target generation process in our CPL method. At the beginning of each training iteration, we first construct a confidence score matrix composed of confidence score vector $\boldsymbol{p}$ for each unlabeled instance. Then, candidate pseudolabels, derived from both intra- and inter-level selection, are extracted to formulate the training target $\boldsymbol{s}$ for the subsequent model training.} 
    \vspace{-1mm}
\label{fig:generate_candidate}
\end{figure}

\textbf{Notations.}
Given an unlabeled example, the set of candidate pseudolabels is denoted by $S$. 
For the unlabeled set $D_{\rm UL} = \{(\boldsymbol{x}_i)\}_{i=1}^N$ composed of $N$ instances, the pair of the instance and the corresponding set of candidate pseudolabels for $D_{\rm UL}$ is represented by $\{(\boldsymbol{x}_i, S_i)\}_{i=1}^N$. Suppose a VLM with learnable parameters $\theta$ is represented by $\boldsymbol{f}_\theta$. Given an instance $\boldsymbol{x}$ in $D_{\rm UL}$, the output of the model can be denoted by $\bs{f}_\theta(\boldsymbol{x})$. Then, we can obtain a vector of confidence scores for all classes via the Softmax function $\bs{g}(\cdot)$, i.e., $\bs{p} = \bs{g}(\bs{f}_\theta(\boldsymbol{x})) = (p_1,p_2,\ldots,p_C)^\top$. 
For convenience, we represent $p_{ic}$ as the confidence score of the $c$-th class for the $i$-th instance. 
By constructing a confidence score matrix $(\bs{p}_1,\bs{p}_2,...,\bs{p}_N)^{\top}$
for all unlabeled data (as shown in Figure \ref{fig:generate_candidate}), we propose to generate candidate pseudolabels by simultanesouly considering two aspects, including both the intra- and inter-instance label selection, respectively.

\textbf{Intra-instance Label Selection.} The concept of intra-instance label selection originates from the idea of selecting the top-$K$ confident labels as the most probable label candidates for each instance.
However, we further consider that it may not be reasonable to select an equal number of top-$K$ confident labels as the candidate set for each instance, because of the varying levels of identification difficulty associated with each instance.

In response to this issue, we propose an adaptive strategy where different values of $K$ are set for instances with different identification difficulty. 
Specifically, for each unlabeled instance $\bs{x}_i$, potential labels are sequentially incorporated into the candidate set $S^{\mathrm{intra}}_i$. This is done in \emph{descending order} based on the corresponding confidence scores until the cumulative confidence score just surpasses a threshold $\tau$. 
Formally, for an unlabeled instance $\bs{x}_i$, its selected candidate set $S^{\text{intra}}_i$ is represented as 
\begin{gather}\label{eq:eq_S_intra}
    S^{\text{intra}}_i = \text{MinSize}(\{c \, | \, \sum\nolimits_{c=1}^C {p_{ic} \geqslant \tau}\}),
\end{gather}
where $\mathrm{MinSize}(\cdot )$ means a function that returns the set with the minimal size and we assume that we select the elements $({p}_{i1}, \ldots, p_{iC})$ from the largest to the smallest.
The hyper-parameter $\tau$ serves as a threshold to ensure that the candidate set $S^{\text{intra}}_i$ for each instance $\bs{x}_i$ encompasses a nearly equivalent level of confidence scores, thereby guaranteeing a comparable level of label estimation accuracy for the candidate pseudolabels of each instance.

As for the determination of the threshold $\tau$, considering the model’s average prediction confidence increases with the training process, it is natural to think that the threshold could also be adaptively updated with the model’s training. Therefore, we obtain $\tau$ from the prediction confidence among $D_{\rm UL}$ and update at the beginning of each training iteration. Specifically, for an unlabeled instance $\bs{x}_i$, the prediction confidence is represented as $\hat{p}_i = \max_{c\in [C]}({p}_{ic})$, which is the maximum value among the confidence scores of instance $\bs{x}_i$.
Subsequently, we can obtain a list of the prediction confidence denoted as $\hat{\bs{p}} = ({\hat p}_1,{\hat p}_2,...,{\hat p}_N) \in [0,1]^N$. By sorting $\hat{\bs{p}}$ in ascending order, $\tau$ is determined when the ratio $\alpha$ is given:
\begin{equation}
\label{eq:tau_adaptive}
    \tau = \text{Quantile}(\text{Sort}(\hat{\bs{p}}), \alpha).
\end{equation}
Here, $\mathrm{Quantile}(\cdot, \alpha)$ is a function that returns the value at the given quantile of the vector and $\mathrm{Sort}(\cdot)$ is a function that sorts the vector in ascending order. $\alpha$ is a hyper-parameter that denotes the specified quantile.
It is noteworthy that, based on the formula in \cref{eq:tau_adaptive}, setting $\alpha=0\%$ would result in an extreme case where all the candidates for pseudolabels contain only the most confident label. This is equivalent to the hard pseudolabeling, further demonstrating the flexibility of our candidate set generation strategy.

\textbf{Inter-instance Label Selection}. Since the predictions from CLIP exhibits class imbalance performance across various categories (see Figure \ref{fig:matrixa}), the generated single hard pseudolabels are typically class-imbalanced. 
To balance the ratio of each category in the candidate pseudolabel set and mitigate the overwhelming influence of dominant classes in pseudolabel generation, we also employ an inter-instance label selection strategy to further refine the candidate pseudolabels. 

Specifically, for each class $c\in [C]$, we employ a vector ${\bs{q}_c} = (p_{1c}, p_{2c}, \ldots, p_{Nc})$ to denote the confidence scores across all $N$ instances (i.e., the column vector of the confidence score matrix illustrated in \cref{fig:generate_candidate}). 
Then, we sort vector $\bs{q}_c$ in ascending order and construct the candidate pseudolabel set $S^{\text{inter}}_i$ of an instance $\bs{x}_i$. Class $c$ is included in $S^{\text{inter}}_i$, when $p_{ic}$ exhibits relatively higher confidence levels within the vector $\bs{q}_c$.
Given a selection ratio $\beta$, for an unlabeled instance $\bs{x}_i$, its candidate pseudolabel set $S^{\text{inter}}_i$ is constructed by
\begin{equation}\label{eq:eq_S_inter}
    S^{\text{inter}}_i =  \{c \mid p_{ic} > \text{Quantile}(\text{Sort}(\bs{q}_c), \beta)\}_{c=1}^C. 
\end{equation}
Eventually, for each unlabeled instance $\bs{x}_i$, its final candidate pseudolabel set $S_i$ is obtained by the intersection of the two candidate pseudolabel sets $S_i^{\mathrm{intra}}$ in Eq.~(\ref{eq:eq_S_intra}) and $S_i^{\mathrm{inter}}$ in Eq.~(\ref{eq:eq_S_inter}). This can be formally expressed as
\begin{equation}
    S_i = S^{\text{intra}}_i \cap S^{\text{inter}}_i. 
\end{equation}
This refined strategy for candidate pseudolabel generation ensures a more balanced and accurate representation of classes, thereby enhancing the model's ability to learn from a diverse and equitable distribution of pseudolabels.

\subsection{Learning with Candidate Pseudolabels}
\label{sec:subsec3.2}
With the generated candidate pseudolabels for unlabeled data, we need to construct a suitable training objective to learn from such supervision information. Fortunately, many loss functions have been designed for learning with candidate pseudolabels (a.k.a. partial-label learning) \cite{feng2020provably,wen2021leveraged,zhang2021exploiting}, which can be directly used even without any modifications.

At the start of each training iteration, we construct the training set, denoted as $D_{\rm T}$, from the unlabeled set $D_{\rm UL}$. This set contains $M$ instance-candidate pairs $\{(\boldsymbol{x}_i, S_i)\}_{i=1}^M$. In simple terms, we filter out instances with no candidate labels in $D_{\rm UL}$ and add the remaining instances, along with their candidate labels, into $D_{\rm T}$. The construction of $D_{\rm T}$ can be represented as
\begin{equation}
     D_{\rm T} = \{(\boldsymbol{x}_i, S_i) \mid |S_i| > 0 \}_{i=1}^N.
\end{equation}
In practical training, the training target of candidate pseudolabels for each unlabeled instance $\bs{x}_i$ can be re-expressed as a vector $\bs{s}_i$ with the size of $C$, where $s_{ic}=1$ if $c\in S$ and $s_{ic}=0$ if $c\notin S$, for $c\in [C]$.
Subsequently, the instances from $D_{\rm T}$ and the corresponding training targets are utilized for training the model.

Depending on the availability of labeled data in downstream tasks, the training objective of our method can be divided into two forms:
\begin{itemize}[leftmargin=3mm]
\setlength\itemsep{0mm}
    \item In the context of semi-supervised learning and transductive zero-shot learning, a small labeled set $D_{\rm L} = \{(\bs{x}_i, \bs{y}_i)\}_{i=1}^O$ is provided. Consequently, at each iteration, we have two sets: labeled set $\{(\bs{x}_i, \bs{y}_i)\}_{i=1}^{b_1}$ with a batch size of $b_{1}$ and $\{(\bs{x}_i, \bs{s}_i)\}_{i=1}^{b_2}$ with a batch size of $b_{2}$. The training objective is
    \begin{align*}\label{eq:labeled}
        \mathcal{L} &= \mathcal{L}_{\rm L} + \lambda \mathcal{L}_{\rm UL} \\
        &= \frac{1}{b_{1}} \sum\nolimits_{i=1}^{b_{1}} {L_{\mathrm{ce}}(\bs
{x}_i, \bs{y}_i) + \lambda
            \frac{1}{b_{2}} \sum\nolimits_{i=1}^{b_{2}} {L_{\mathrm{p}}(\bs{x}_i, {\bs{s}}_i)}},
    \end{align*}
where $L_{\mathrm{ce}}$ represents the cross-entropy loss function used for learning with the labeled set, and $L_\mathrm{p}$ denotes the loss function used for learning with the candidate pseudolabel set. During training, we typically set $b_{2}$ to a pre-fixed value. For $b_{1}$, we set it as $({|D_{\rm L}|}/{|D_{\rm T}|})\times b_{2}$, ensuring that $D_{\rm L}$ and $D_{\rm T}$ have similar iteration counts throughout the training process.

\item In the unsupervised learning setting, we only have access to the unlabeled data $D_{\rm UL}$. Given each mini-batch of training data $\{(\bs{x}_i, {\bs{s}}_i)\}_{i=1}^{b_{2}}$, the training objective is
    \begin{equation}\label{eq:unlabeled}
        \mathcal{L} =  \mathcal{L}_{\rm UL} 
        = \frac{1}{b_{2}} \sum\nolimits_{i=1}^{b_{2}} { L_{\mathrm{p}}(\bs{x}_i, {\bs{s}}_i)}. \nonumber
    \end{equation}
\end{itemize}

\textbf{Update of Candiate Pseudolabels}. 
Throughout the training process, the candidate pseudolabels are progressively updated after each pre-defined iteration. For each iteration, we regenerate the candidate pseudolabels for all unlabeled data based on intra- and inter-instance label selection, learning with the newly generated candidate pseudolabels
The detailed iterative process can be found in Appendix \ref{sec:appendix_a}.

\section{Experiments}
\label{sec:Experiments}
\begin{table*}[t]
    \centering
    \caption{
    Comparison results of top-1 test accuracy (\%) on six benchmarks when applying \textbf{Textual prompts} as tuning strategy. Note that ``\ding{51}" and ``\ding{55}" denote whether full unlabeled data are utilized for fine-tuning or not, respectively.
    }
    \vspace{-2mm}
        \resizebox{0.98\linewidth}{!}{
    \begin{tabular}{lc|ccccccccc}
    \toprule[1.1pt]
        &   & \multicolumn{3}{c}{Flowers102} & \multicolumn{3}{c}{RESISC45} & \multicolumn{3}{c}{DTD} \\ \cmidrule{3-11}
        \multicolumn{2}{c|}{Methods}     & SSL & UL & TRZSL & SSL & UL & TRZSL & SSL & UL & TRZSL \\
    \cmidrule(lr){1-2} \cmidrule(lr){3-5} \cmidrule(lr){6-8} \cmidrule(lr){9-11} 
    Zero-shot CLIP & & \multicolumn{2}{c}{$63.67_{0.00}$} & $63.40_{0.00}$ & \multicolumn{2}{c}{$54.48_{0.00}$} & $54.46_{0.00}$ &  \multicolumn{2}{c}{$43.24_{0.00}$} & $43.45_{0.00}$ \\
    FPL & \ding{55} &  $75.96_{0.74}$ & $65.67_{0.23}$ & $80.97_{0.00}$ & $68.13_{0.55}$ & $63.07_{0.38}$ & $72.11_{0.00}$ & $37.10_{5.45}$ & $44.96_{0.55}$ & $46.30_{0.03}$ \\
    \cellcolor{Gray}\ourmethodCC {\small (Ours)} & \cellcolor{Gray}\ding{55}   &  \cellcolor{Gray}$\textbf{77.36}_{0.24}$ & \cellcolor{Gray}$\textbf{70.01}_{0.21}$ & \cellcolor{Gray}$\textbf{84.60}_{0.10}$ & \cellcolor{Gray}$\textbf{71.73}_{0.57}$ & \cellcolor{Gray}$\textbf{
68.47}_{0.34}$ & \cellcolor{Gray}$\textbf{72.16}_{0.26}$ & \cellcolor{Gray}$\textbf{54.63}_{0.79}$ & \cellcolor{Gray}$\textbf{48.92}_{0.17}$ & \cellcolor{Gray}$\textbf{59.79}_{1.32}$
 \\
    \midrule
    GRIP & \ding{51} & $ {83.60}_{0.48}$ & $ {69.84}_{1.06}$ & $ {86.26}_{0.00}$ & $ {74.11}_{0.68}$ & $ {70.55}_{0.88}$ & $ {81.07}_{0.00}$ & $ {56.07}_{0.85}$ & $ {46.09}_{1.06}$ & $ {65.30}_{0.01}$  \\
     \cellcolor{Gray}\ourmethodCC {\small (Ours)} &  \cellcolor{Gray}\ding{51} &  \cellcolor{Gray}$\textbf{89.66}_{0.36}$ &  \cellcolor{Gray}$\textbf{72.90}_{0.78}$ &  \cellcolor{Gray}$\textbf{87.35}_{0.76}$ &  \cellcolor{Gray}$\textbf{80.98
}_{0.11}$ &  \cellcolor{Gray}$\textbf{77.39}_{0.44}$ &  \cellcolor{Gray}$\textbf{85.85}_{0.49}$ &  \cellcolor{Gray}$\textbf{61.21}_{0.56}$ &  \cellcolor{Gray}$\textbf{51.91}_{0.71}$ &  \cellcolor{Gray}$\textbf{68.00}_{0.34}$    \\
    \midrule\midrule
        &   & \multicolumn{3}{c}{CUB} & \multicolumn{3}{c}{EuroSAT} & \multicolumn{3}{c}{FGVCAircraft} \\ \cmidrule{3-11}
    \multicolumn{2}{c|}{Methods}     & SSL & UL & TRZSL & SSL & UL & TRZSL & SSL & UL & TRZSL \\
    \cmidrule(lr){1-2} \cmidrule(lr){3-5} \cmidrule(lr){6-8} \cmidrule(lr){9-11} 
    Zero-shot CLIP & & \multicolumn{2}{c}{$51.82_{0.00}$} & $51.57_{0.00}$ & \multicolumn{2}{c}{$32.88_{0.00}$} & $30.54_{0.00}$ &  \multicolumn{2}{c}{$ {17.58}_{0.00}$} & $17.86_{0.00}$ \\
    FPL & \ding{55} &  $55.29_{0.59}$ & $53.04_{0.53}$ & $55.44_{0.20}$ & $ {62.05}_{1.64}$ & $48.96_{1.49}$ & $53.70_{26.87}$ & $ {20.02}_{0.77}$ & $ {16.62}_{0.67}$ & $17.55_{0.37}$ \\
    \cellcolor{Gray}\ourmethodCC {\small (Ours)} & \cellcolor{Gray}\ding{55}&  
    \cellcolor{Gray}$\textbf{56.37}_{0.45}$ & \cellcolor{Gray}$\textbf{54.18}_{0.05}$ & \cellcolor{Gray}$\textbf{64.01}_{0.17}$ & \cellcolor{Gray}$\textbf{64.84}_{2.15}$ & \cellcolor{Gray}$\textbf{51.45}_{1.97}$ & \cellcolor{Gray}$\textbf{54.03}_{2.27}$ & \cellcolor{Gray}$\textbf{22.37}_{0.66}$ & \cellcolor{Gray}$\textbf{18.90}_{0.20}$ & \cellcolor{Gray}$\textbf{28.47}_{0.43}$ \\
    \midrule
    GRIP & \ding{51} &  $ {56.65}_{0.33}$ & $ {51.42}_{0.21}$ & $ {59.48}_{0.38}$ & $58.66_{2.64}$ & $ {57.21}_{1.77}$ & $ {92.33}_{0.69}$ & $16.98_{0.82}$ & $15.22_{0.71}$ & $ {26.08}_{0.25}$\\
   \cellcolor{Gray}\ourmethodCC {\small (Ours)} & \cellcolor{Gray}\ding{51}&  
    \cellcolor{Gray}$\textbf{58.53}_{0.24}$ & \cellcolor{Gray}$\textbf{53.47}_{0.36}$ & \cellcolor{Gray}$\textbf{66.20}_{0.50}$ & \cellcolor{Gray}$\textbf{77.51}_{0.80}$ & \cellcolor{Gray}$\textbf{67.26}_{0.47}$ & \cellcolor{Gray}$\textbf{93.78}_{0.12}$ & \cellcolor{Gray}$\textbf{22.48}_{0.63}$ & \cellcolor{Gray}$\textbf{18.35}_{0.27}$ & \cellcolor{Gray}$\textbf{30.86}_{0.70}$ \\
    \bottomrule[1.1pt]
    \end{tabular}}
\label{tab:textul_tuning}
\end{table*}

\begin{table*}[t]
    \caption{
    Comparison results of top-1 test accuracy (\%) on six benchmarks when applying \textbf{Visual prompts} as tuning strategy. Note that ``\ding{51}" and ``\ding{55}" denote whether full unlabeled data are utilized for fine-tuning or not, respectively.
    }
    \vspace{-2mm}
    \centering
        \resizebox{0.98\linewidth}{!}{
    \begin{tabular}{lc|ccccccccc}
    \toprule[1.1pt]
        &   & \multicolumn{3}{c}{Flowers102} & \multicolumn{3}{c}{RESISC45} & \multicolumn{3}{c}{DTD} \\ \cmidrule{3-11}
        \multicolumn{2}{c|}{Methods}     & SSL & UL & TRZSL & SSL & UL & TRZSL & SSL & UL & TRZSL \\
    \cmidrule(lr){1-2} \cmidrule(lr){3-5} \cmidrule(lr){6-8} \cmidrule(lr){9-11} 
    Zero-shot CLIP & & \multicolumn{2}{c}{$63.67_{0.00}$} & $63.40_{0.00}$ & \multicolumn{2}{c}{$54.48_{0.00}$} & $54.46_{0.00}$ &  \multicolumn{2}{c}{$43.24_{0.00}$} & $43.45_{0.00}$ \\
    FPL  & \ding{55} &  $67.03_{0.65}$ & $ {65.50}_{0.41}$ & $71.94_{0.00}$ & $65.14_{0.25}$ & $62.24_{0.22}$ & $67.85_{0.00}$ & $47.60_{1.09}$ & $47.69_{0.48}$ & $52.43_{0.00}$  \\
    \cellcolor{Gray}\ourmethodCC {\small (Ours)} & \cellcolor{Gray}\ding{55} 
    & \cellcolor{Gray}$\textbf{70.58}_{0.13}$ & \cellcolor{Gray}$\textbf{68.94}_{0.16}$ & \cellcolor{Gray}$\textbf{78.13}_{0.31}$ & \cellcolor{Gray}$\textbf{68.85}_{0.13}$ & \cellcolor{Gray}$\textbf{67.97}_{0.17}$ & \cellcolor{Gray}$\textbf{72.18}_{0.27}$ & \cellcolor{Gray}$\textbf{52.64}_{0.68}$ & \cellcolor{Gray}$\textbf{50.37}_{0.46}$ & \cellcolor{Gray}$\textbf{55.90}_{0.69}$\\ \midrule 
    GRIP  & \ding{51} &  $ {67.95}_{1.2}$ & $63.09_{0.56}$ & $ {77.18}_{0.00}$ & $ {71.22}_{0.77}$ & $ {68.43}_{0.61}$ & $ {82.19}_{0.00}$ & $ {54.57}_{4.86}$ & $ {50.51}_{0.99}$ & $ {62.78}_{0.00}$ \\
    \cellcolor{Gray}\ourmethodCC {\small (Ours)} & \cellcolor{Gray}\ding{51} 
    & \cellcolor{Gray} $\textbf{73.52}_{0.37}$ & \cellcolor{Gray}$\textbf{67.25}_{0.41}$ & \cellcolor{Gray}$\textbf{80.14}_{0.73}$ & \cellcolor{Gray}$\textbf{78.46}_{0.74}$ & \cellcolor{Gray}$\textbf{72.97}_{0.58}$ & \cellcolor{Gray}$\textbf{86.67}_{0.33}$ & \cellcolor{Gray}$\textbf{58.74}_{0.81}$ & \cellcolor{Gray}$\textbf{53.42}_{0.56}$ & \cellcolor{Gray}$\textbf{65.31}_{0.78}$ \\
    \midrule\midrule
        &   & \multicolumn{3}{c}{CUB} & \multicolumn{3}{c}{EuroSAT} & \multicolumn{3}{c}{FGVCAircraft} \\ \cmidrule{3-11}
    \multicolumn{2}{c|}{Methods}     & SSL & UL & TRZSL & SSL & UL & TRZSL & SSL & UL & TRZSL \\
    \cmidrule(lr){1-2} \cmidrule(lr){3-5} \cmidrule(lr){6-8} \cmidrule(lr){9-11} 
    Zero-shot CLIP & & \multicolumn{2}{c}{$51.82_{0.00}$} & $51.57_{0.00}$ & \multicolumn{2}{c}{$32.88_{0.00}$} & $30.54_{0.00}$ &  \multicolumn{2}{c}{$ {17.58}_{0.00}$} & $17.86_{0.00}$ \\
    FPL & \ding{55}& $52.86_{0.42}$  & $53.17_{0.06}$ &  $54.17_{0.16}$ & $52.47_{2.53}$ & $48.79_{3.69}$ & $68.68_{14.74}$ & $ {20.14}_{0.26}$ & $ {18.28}_{0.33}$ & $16.28_{0.45}$ \\
    \cellcolor{Gray}\ourmethodCC {\small (Ours)}
    &\cellcolor{Gray}\ding{55}
    &\cellcolor{Gray}$\textbf{53.37}_{0.55}$ 
    &\cellcolor{Gray}$\textbf{53.28}_{0.31}$ 
    &\cellcolor{Gray}$\textbf{56.43}_{0.21}$ & \cellcolor{Gray}$\textbf{66.37}_{2.10}$ & \cellcolor{Gray}$\textbf{52.83}_{2.10}$ & \cellcolor{Gray}$\textbf{74.02}_{1.34}$ & \cellcolor{Gray}$\textbf{21.52}_{0.68}$ & \cellcolor{Gray}$\textbf{20.10}_{0.51}$ & \cellcolor{Gray}$\textbf{26.73}_{0.08}$ \\ \midrule
    GRIP & \ding{51} &  $ \textbf{{53.83}}_{0.11}$ & $ \textbf{{52.91}}_{0.26}$ & $ {54.92}_{0.17}$ & $ {63.48}_{3.09}$ & $ {63.68}_{3.42}$ & $ {96.97}_{0.77}$ & $ {19.43}_{0.50}$ & $ {17.51}_{0.61}$ & $ {26.42}_{0.30}$ \\
    \cellcolor{Gray}\ourmethodCC {\small (Ours)} & \cellcolor{Gray}\ding{51}
    & \cellcolor{Gray}$49.50_{0.42}$ & \cellcolor{Gray}$52.11_{0.24}$ & \cellcolor{Gray}$\textbf{56.37}_{0.06}$ & \cellcolor{Gray}$\textbf{72.03}_{1.24}$ & \cellcolor{Gray}$\textbf{68.93}_{1.15}$ & \cellcolor{Gray}$\textbf{98.31}_{0.18}$ & \cellcolor{Gray}$\textbf{20.51}_{0.68}$ & \cellcolor{Gray}$\textbf{18.26}_{0.38}$ & \cellcolor{Gray}$\textbf{30.26}_{0.46}$ \\
        \bottomrule[1.1pt]
    \end{tabular}}

\label{tab:visual_prompt}
\end{table*}

To evaluate the effectiveness of our proposed candidate pseudolabel learning (CPL) method, we implement experiments in several dimensions. \ding{182} \textbf{Learning paradigm variety}: we conduct experiments on three learning paradigms including semi-supervised learning, unsupervised learning, and transductive zero-shot learning. \ding{183} \textbf{Prompt tuning variety}: all methods are tested with a textual prompt as well as a visual prompt as CLIP's learnable parameters and tuning strategy. \ding{184} \textbf{Task variety}: we evaluate the effectiveness of CPL on nine classification tasks.

\subsection{Experimental Setting}
\label{sec4.1:exp_setting}
\textbf{Datasets.}
We conduct an extensive evaluation of our method on nine classification datasets from diverse domains, including FGVC-Aircraft \cite{maji2013fine}, EuroSAT \cite{helber2019eurosat}, CUB \cite{WahCUB_200_2011}, Flowers102 \cite{nilsback2008automated}, RESISC45 \cite{cheng2017remote}, DTD \cite{cimpoi2014describing}, CALTECH-101 \cite{fei2004learning}, UCF-101 \cite{soomro2012ucf101}, and CIFAR-100 \cite{krizhevsky2009learning}. 

\textbf{Learning Paradigms.} To comprehensively evaluate the performance of our method, we consider three common scenarios involving the use of unlabeled data:  Unsupervised Learning (UL), Semi-Supervised Learning (SSL), and Transductive Zero-Shot Learning (TRZSL) tasks. The details of each paradigm and how labeled data issues are handled can be found in Appendix \ref{sec:appendix_c1}. 

\textbf{Hyper-parameter Configuration.} Unless otherwise specified, our experiments utilize ViT-B/32 as the visual backbone, and the prefix size is set at 16 for both textual and visual prompt learning. Also, we adopt the Classifier-Consistent (CC) \cite{feng2020provably} as the default loss function for learning with candidate labels in our method. The default prompt ``\texttt{a photo of a [CLASS]}” is employed to obtain initial predictions from CLIP on all unlabeled instances. 
We adopt SGD as the optimizer and conduct training for 50 epochs. The learning rate is set at 0.0001 for two warm-up cycles, after which it is adjusted to 0.02 and decays following the cosine annealing rule. For SSL and TRZSL, we just set $\lambda$ to 1. 
Regarding candidate pseudolabel update, we designate the iteration number for the CUB dataset as $T=5$ and $T=10$ for all other datasets. Further technical specifics can be found in Appendix \ref{sec:appendix_c2}.

\subsection{Comparison with Previous Methods}
\textbf{Experimental Design and Baselines}. We carry out experiments on two fine-tuning scenarios: tuning with few-shot unlabeled data and tuning with full unlabeled data. 
In the former scenario, we select $q$ samples per class from all unlabeled data, making it more appropriate for rapid fine-tuning on the downstream dataset. In our experiments, we set $q=16$ for all methods to ensure fair comparisons. 
By contrast, the latter scenario fine-tunes CLIP on the entire unlabeled data to achieve superior performance. 
We compare our CPL with two existing methods, namely, Few Pseudolabels (FPL)~\cite{menghini2023enhancing} and Grow and Refine Iteratively Pseudolabels (GRIP)~\cite{menghini2023enhancing}, across six classification tasks under these two scenarios. 
The performance of each method is reported by calculating the test set accuracy, averaged over three runs. For TRZSL, we report the harmonic mean of the accuracies of seen and unseen classes.

Results about textual prompt tuning and visual prompt tuning are shown in \cref{tab:textul_tuning} and \ref{tab:visual_prompt}, respectively. We observed that: 
\ding{182} \textit{Our method consistently outperforms existing hard pseudolabel methods.} Our proposed CPL framework consistently surpasses FPL and GRIP, across a variety of tasks and datasets under both scenarios of tuning with few-shot unlabeled data and tuning with full unlabeled data. This underscores the efficacy of our approach in enhancing the performance of CLIP in downstream tasks. 
\ding{183} \textit{Our method exhibits less dependency on the accuracy of zero-shot CLIP and utilizes unlabeled data more effectively.} Our method excels in settings where the zero-shot capability of CLIP is relatively low and the labeled data is scarce. As shown in Table \ref{tab:textul_tuning}, our method performs well on the EuroSAT dataset, where the initial performance of CLIP is subpar. When tuning with the full unlabeled data, our method improves the top-1 test accuracy by 18.85\% and 10.05\% in the SSL and UL paradigms, compared with GRIP. Similarly, on the DTD dataset, our method improves the top-1 test accuracy by 5.14\% and 5.82\% in the SSL and UL paradigms, compared with GRIP. 
These results demonstrate that our method has a lower requirement for the initial zero-shot capability, rendering it more robust across various learning paradigms and tasks.

\begin{table*}[htb]
\setlength\tabcolsep{10pt}
    \small
    \caption{Comparison results of top-1 test accuracy (\%) on unsupervised learning when applying \textbf{parameter-efficient tuning}. The best and second-best performances are highlighted via \textbf{bold} and \underline{underline}, respectively.} 
    \vspace{-2mm}
    \label{tab:versatility}
    \centering
    \begin{tabular}{l|c|c|c|c|c|c}
    \toprule[0.9pt]
         & Flowers-102 & UCF-101 & CIFAR-100 & EuroSAT & DTD & CALTECH-101\\
\midrule
   CLIP &             {$66.6$} &             {$61.0$} &             {$64.2$} &             {$45.1$} &             {$42.9$} &               {$90.5$} \\
CLIP-PR &             {$57.7$} &             {$57.9$} &             {$63.2$} &             {$44.2$} &             {$40.1$} &               {$84.8$} \\
    UPL &             {\underline{$71.5$}} &             {$63.9$} &             {$65.8$} & {{$62.2$}} &  {\underline{$48.0$}} &            {$90.6$} \\
\SOTAmethod &  {$71.0$} &  {\underline{$68.2$}} &  {\underline{$74.6$}} &  {\underline{$73.9$}} & {$46.1$} &    {\underline{$93.3$}} \\
    \midrule
 \cellcolor{Gray}LaFTer + Ours 
&\cellcolor{Gray}{$\textbf{76.7}$} 
&\cellcolor{Gray}{$\textbf{71.0}$} 
&\cellcolor{Gray}{$\textbf{77.3}$} 
&\cellcolor{Gray}{$\textbf{82.2}$} 
&\cellcolor{Gray}{$\textbf{56.3}$} 
&\cellcolor{Gray}{$\textbf{93.4}$} \\
    \bottomrule[0.9pt]
    \end{tabular}
    \vspace{-3mm}
\end{table*}

\subsection{More Analyses}

\textbf{Versatility}.
Our proposed CPL can be universally applied to existing label-free CLIP fine-tuning scenarios, thereby enhancing their performance. To demonstrate this, we substitute the corresponding pseudolabel module in the existing state-of-the-art method with the candidate pseudolabels generation and update module from CPL. We report the performance of the existing four methods (LaFTer \cite{mirza2023lafter}, UPL \cite{huang2022unsupervised}, and CLIP-PR \cite{kahana2022improving}) under a parameter-efficient tuning strategy while incorporating our proposal into LaFTer.

The results are presented in Table \ref{tab:versatility}. The combiner (LaFTer + Ours) consistently performs best across all six benchmarks. On the DTD dataset, an improvement of 10.2\% in top-1 test accuracy is observed compared with the second-best method, LaFTer. 
Significant improvements across these datasets demonstrate the versatility of our proposed candidate pseudolabels. Consequently, our method is not only effective in its own right but can also enhance the performance of existing CLIP fine-tuning methods when integrated.

\begin{table}[t]   
    \centering
    \small
    \caption{Comparison results of top-1 test accuracy (\%) on DTD with textual prompts tuning. The performance of CPL with five different loss functions is evaluated on three tasks.
    }
    \vspace{-2mm}
     \resizebox{0.97\linewidth}{!}{
    \begin{tabular}{lc|ccc}
    \toprule[0.9pt]
    \multicolumn{2}{c|}{Methods}   & SSL & UL & TRZSL \\\midrule
    Zero-shot CLIP                 &  & \multicolumn{2}{c}{$43.24_{0.00}$} & $43.45_{0.00}$  \\ \midrule
    FPL                             & \ding{55} & $37.10_{5.45}$  & $44.96_{0.55}$ & $46.30_{0.03}$\\
    \ourmethodCC$_{\mathrm{Soft\,CE}}$ & \ding{55} & $51.83_{0.62}$ & $47.02_{0.37}$   & $59.69_{0.59}$ \\
    \ourmethodCC$_{\mathrm{CC}}$             & \ding{55} & $54.63_{0.79}$ & $48.92_{0.17}$ & $59.79_{1.32}$ \\
    \ourmethodCC$_{\mathrm{RC}}$             & \ding{55} & $54.98_{0.49}$ & $49.96_
{0.15}$ & $59.42_{0.44}$\\
    \ourmethodCC$_{\mathrm{CAV}}$            & \ding{55} & $55.50_{0.29}$ & $48.69_{0.66}$ & $59.44_{0.13}$\\
    \ourmethodCC$_{\mathrm{LW}}$             & \ding{55} & $55.21_{0.74}$ & $49.82_{0.91}$ & $59.24_{0.72}$\\ \midrule
    GRIP                            & \ding{51} & $56.07_{0.85}$ & $46.09_{1.06}$ & $65.30_{0.01}$  \\
    \ourmethodCC$_{\mathrm{Soft\,CE}}$ & \ding{51} & $60.83_{0.66}$ & $49.13_{0.10}$ & $66.26_{0.77}$ \\
    \ourmethodCC$_{\mathrm{CC}}$             & \ding{51} & $61.21_{0.56}$ & $51.91_{0.71}$ & $68.00_{0.34}$\\
    \ourmethodCC$_{\mathrm{RC}}$             & \ding{51} & $60.21_{0.46}$ & $51.58_{0.11}$ & $67.95_{0.31}$\\
    \ourmethodCC$_{\mathrm{CAV}}$            & \ding{51} & $61.06_{0.50}$ & $49.31_{0.19}$ & $67.76_{0.53}$\\
    \ourmethodCC$_{\mathrm{LW}}$             & \ding{51} & $60.20_{0.69}$ & $52.23_{0.84}$ & $68.29_{0.99}$\\
     \bottomrule[0.9pt]
    \end{tabular}}

\label{tab:various_loss}
\end{table}

\textbf{Training Loss}. Our method is not confined to a particular loss function. Due to this flexibility, it can achieve competitive performance with various loss functions. In addition to the Classifier-Consistent (CC) \cite{feng2020provably}, we have also explored four other loss functions capable of handling learning with multiple candidate labels. These include RC \cite{feng2020provably}, CAV \cite{zhang2021exploiting}, LW \cite{wen2021leveraged}, and a soft target cross-entropy. In our proposed soft target cross-entropy (Soft CE), normalized confidence scores from the model prediction are used as the soft targets for the next iteration. 
We defer the detailed discussion of this soft target scheme to Appendix \ref{sec:appendix_b}.

The comparison results are detailed in Table \ref{tab:various_loss}. We observe that all explored loss functions exhibit competitive performance, significantly outperforming the hard pseudolabel methods, FPL and GRIP. This suggests that our CPL is not overly sensitive to the choice of loss function, further highlighting its flexibility. Moreover, we find that all candidate pseudolabel schemes consistently outperform Soft CE, incorporating more prediction information from the prior model. 
This indicates that the strategy of treating all candidate labels equally can more effectively mitigate the influence of prior category bias, thereby enhancing the performance of CLIP in downstream tasks.

\textbf{Imbalance Dataset}. 
Considering the universality of the class-imbalanced training set underlying unlabeled data, we conduct relevant experiments to evaluate the performance of our proposed method in the class imbalance setting. Specifically, we keep the convention from Zhou et al. \cite{zhou2020bbn} and manually construct an imbalanced CIFAR-100 via an imbalanced ratio $\delta$. We set $\delta=50$ and $\delta=100$ (a larger value of $\delta$ denotes more imbalanced) to compared the performance of CPL and LaFTer on the imbalanced dataset. 
The comparison results are shown in Table \ref{tab:comp_on_imbalance}. 
While class imbalance does exert some influence on the performance of CPL, our results indicate that CPL still outperforms LaFTer, which employs hard pseudolabels, under these imbalanced conditions. 

\begin{table}[th]
    \centering
    \small
    \caption{Comparison of top-1 test accuracy (\%) on CIFAR100 dataset with both balanced and imbalanced distribution. Note that ``w/o inter" denotes the variant of CPL without the inter-instance label selection.}
    \label{tab:comp_on_imbalance}
    \begin{tabular}{l|ccc}
    \toprule[0.9pt]
    Methods & Balanced & \makecell[c]{Imbalanced\\$\delta$=100} & \makecell[c]{Imbalanced\\$\delta$=50} \\
    \midrule
    LaFTer & $74.64$ & $65.63$ & $66.59$ \\
    CPL (w/o inter) & $76.07$ & $66.68$ & $67.85$ \\
    CPL & $\textbf{77.32}$ & $\textbf{67.70}$ & $\textbf{69.65}$ \\
    \bottomrule[0.9pt]
    \end{tabular}
    \vspace{-1mm}
\end{table}

We further explore the impact on CPL's performance when the inter-instance label selection is not applied on imbalanced CIFAR100. As shown in Table \ref{tab:comp_on_imbalance}, the performance of CPL without the inter-instance label selection on imbalanced CIFAR100 is slightly inferior to that of CPL with the inter-instance label selection, especially when the balance ratio is 50. This result further substantiates the necessity of the inter-instance label selection strategy and intra-instance label selection strategy.

\begin{figure}[t]
    \centering     
    \includegraphics[width=0.99\linewidth]{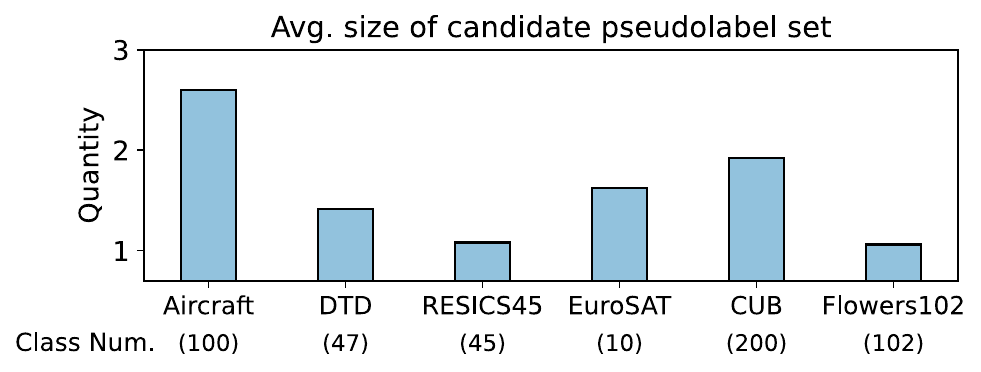}
    \caption{Visualization of the average set size of candidate pseudolabels among all unlabeled data on six datasets under the UL setting of textual prompt tuning.}
    \label{fig:set_size}
    \vspace{-2mm}
\end{figure}

\textbf{Set Size of Candidate Pseudolabels}. 
While increasing the size of the candidate set enhances the likelihood of encompassing the true label, it simultaneously amplifies the ambiguity of the training targets, thereby enhancing the difficulty for the model to learn from the candidate labels. 
We visualize the average size of the generated candidate pseudolabels among all unlabeled data before the last iteration and present the result in Figure \ref
{fig:set_size}. It can be observed that the average size of candidate pseudolabels on the majority of the six datasets is close to 1. This implicitly indicates that a large number of unlabeled data have only one candidate label. Consequently, the candidate pseudolabels in our method would not introduce a high degree of ambiguity or high entropy optimization objectives.

\textbf{Different Proportion of Unlabeled Data}. 
When a large amount of unlabeled downstream data is available, efficiently leveraging it under resource constraints becomes crucial. Typically, utilizing more unlabeled data can yield improved performance but at the cost of increased computational requirements and training time. The question then arises: is it more beneficial to invest additional resources to use all the unlabeled data, or is it better to train with a small amount of high-quality data? To explore this, we compared the performance improvement of CPL on different datasets when a certain proportion of unlabeled data or a small amount of well-labeled data is used for training. 

\begin{figure}[t]
    \centering     
    \includegraphics[width=0.999\linewidth]{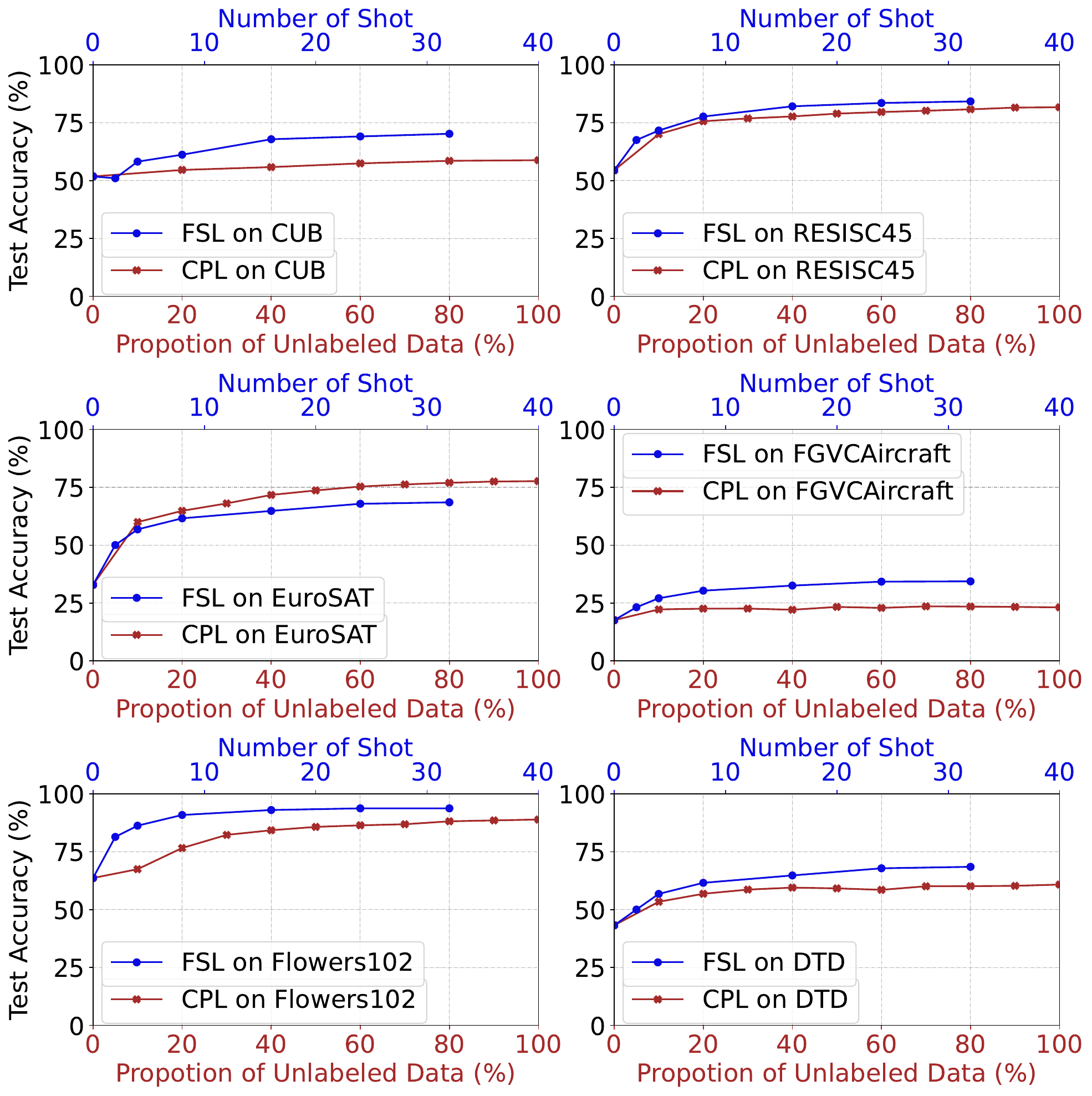}
    \caption{Visualization of the performance improvement of CLIP with CPL (each class only has two labeled data) and fully supervised few-shot learning when textual prompt tuning is applied. The $x$-axis in {\color{blue}blue} represents the number of labeled instances, while the $x$-axis in {\color{bred}red} represents the proportion of the unlabeled dataset. Both lines originate from the zero-shot performance of CLIP.}
    \label{fig:compare_saturation}
    \vspace{-2mm}
\end{figure}

In Figure \ref{fig:compare_saturation}, a common trend is observed across all datasets: as the proportion of data used for training increases, the performance improvement of CPL gradually diminishes, eventually reaching a saturation point. Simultaneously, we also find that few-shot learning usually achieves a higher performance improvement than methods primarily relying on unlabeled data, even though the quantity of used unlabeled data is larger. 

This is particularly significant on fine-grained datasets with a large number of classes, such as CUB and FGVC-Aircraft. In addition, the onset of performance saturation when using unlabeled data also happens earlier on these datasets. These observations suggest that when a downstream dataset is challenging, using more well-labeled data for training may be a more efficient way to improve the performance of CLIP under resource constraints.

\begin{figure}[t]
    \centering     
    \includegraphics[width=0.99\linewidth]{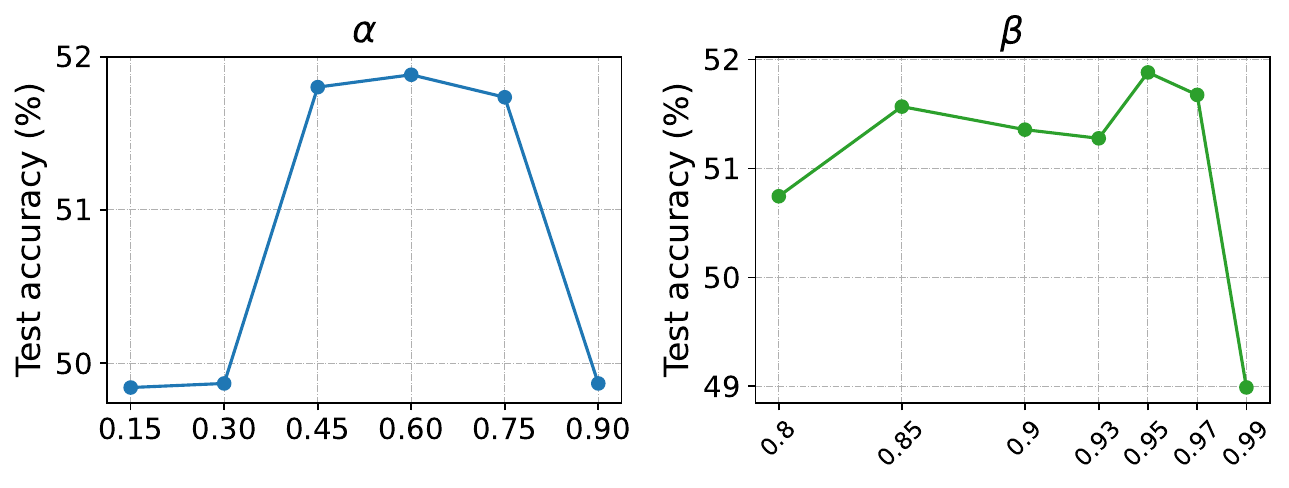}
    \caption{Hyperparameters evaluation on DTD dataset under the UL setting. We illustrate the performance of our method while keeping $\beta=0.95$ (left) and $\alpha=0.60$ (right) constant.}
    \label{fig:hyperparameter}
\end{figure}

\subsection{Ablation Studies} 

\textbf{Sensitivity Analysis.}
There are mainly two hyperparameters in our CPL method, containing $\alpha$ in Eq. (\ref{eq:tau_adaptive}) and $\beta$ in Eq. (\ref{eq:eq_S_inter}). We conduct a sensitivity study to explore the impact of each hyperparameter on CPL's performance. In Figure \ref{fig:hyperparameter}, we plot the performance of one hyperparameter while keeping the other constant. 
Generally, the larger the value of $\alpha$ and the smaller the value of $\beta$, the more candidate pseudolabels CPL tends to generate during the pseudolabel generation process. 
In this figure, we find that maintaining the value of $\alpha$ between 0.45 and 0.75 and $\beta$ between 0.95 and 0.97 yields better performance. However, when $\beta$ exceeds 0.99, the performance declines as CPL tends to generate fewer candidate pseudolabels. Consequently, a large number of unlabeled samples end up with an empty set of candidate labels, reducing the utilization of unlabeled samples. 
More analyses about these two hyperparameters via grid search are shown in Appendix \ref{sec:appendix_D}. 

\begin{figure}[t]
\centering     
\includegraphics[width=0.999\linewidth]{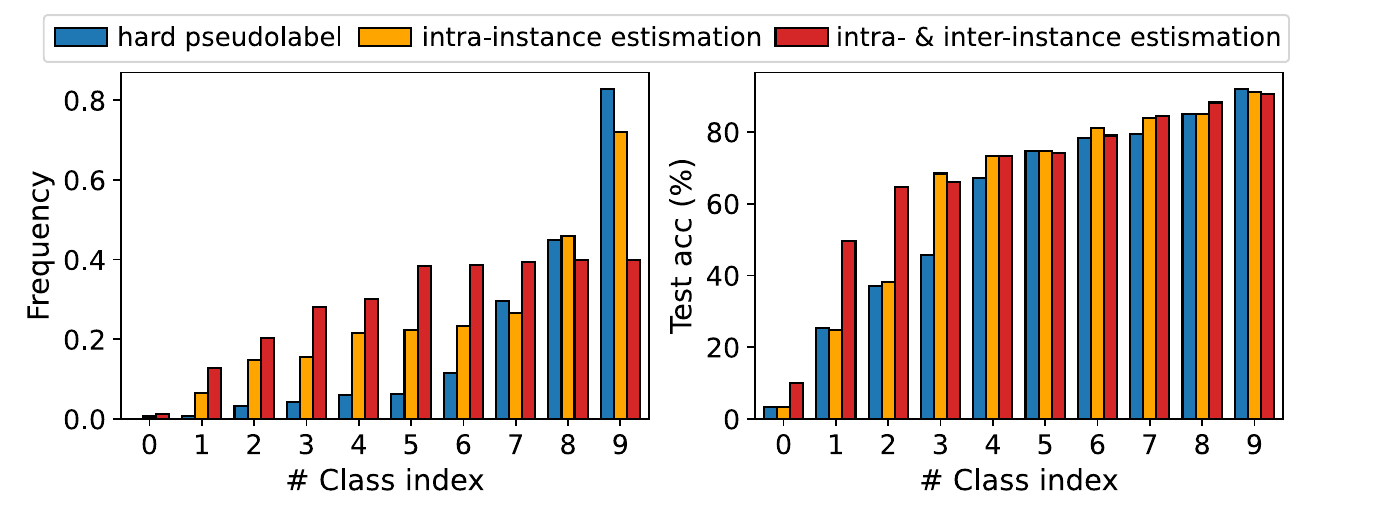}
\caption{\emph{left}: The frequency distribution of the generated pseudolabels with different strategies. \emph{right}: The influence of each strategy on the class-level test accuracy for EuroSAT.}
\vspace{-2mm}
\label{fig:class_frequency}
\end{figure}

\textbf{Effect of Each Selection Strategy.}
The core of our proposed CPL is the generation of candidate pseudolabels, which includes two label selection strategies. We evaluate the effectiveness of each strategy using two metrics: frequency of different classes in pseudolabels and class-level test accuracy.
Firstly, compared to hard pseudolabels, the intra-instance label selection strategy results in a more class-balanced training target (as shown on the left of Figure \ref{fig:class_frequency}), leading to improved test accuracy. 
Furthermore, incorporating an additional inter-instance label selection enables CPL to achieve a more balanced distribution among different classes, particularly evident in classes 5-9 as shown on the left of Figure \ref{fig:class_frequency}. 
The right side of Figure \ref{fig:class_frequency} illustrates the impact of each strategy on the class-level test accuracy. The results indicate that both strategies contribute to the improvement in test accuracy.
These results underscore two key points: \ding{182} both intra-instance and inter-instance label selection strategies are crucial for generating high-quality candidate pseudolabels and \ding{183} the inter-instance label selection can further enhance class balance and test accuracy.

\textbf{Different Image Encoders.}
We further conduct experiments to evaluate the effect of different image encoders. The comparison results on Flowers102, RESISC45, and DTD using ViT-L/14 are presented in Table \ref{tab:diff_backbone}. Our method consistently surpasses the previous methods, demonstrating the effectiveness of our approach in enhancing the performance of CLIP in downstream tasks when larger image encoders are employed. This suggests that our method is capable of better leveraging unlabeled data to improve the performance of CLIP, even when the model size is increased.

\begin{table}[t]   
    \centering
    \small
    \caption{Comparison results of a different image encoder (ViT-L/14) when applying textual prompt tuning.}
    \vspace{-2mm}
     \resizebox{0.99\linewidth}{!}{
    \begin{tabular}{clc|ccc}
    \toprule[0.9pt]
    \multicolumn{3}{c|}{Methods}   & SSL & UL & TRZSL \\\midrule
   \multirow{5}{*}{\rotatebox{90}{DTD}} &Zero-shot CLIP  &  & \multicolumn{2}{c}{$52.45_{0.00}$} & $51.61_{0.00}$  \\
    &FPL                           & \ding{55} & $60.61_{1.56}$ & $52.99_{0.43}$ & $60.77_{0.54}$  \\
    &\cellcolor{Gray}\ourmethodCC     & \cellcolor{Gray}\ding{55} & \cellcolor{Gray}$\textbf{62.78}_{0.17}$ & \cellcolor{Gray}$\textbf{57.23}_{0.19}$ & \cellcolor{Gray}$\textbf{62.52}_{1.38}$\\     
    &GRIP                             & \ding{51} & $60.91_{0.00}$ & $54.40_{0.00}$ & $64.92_{0.00}$  \\
    &\cellcolor{Gray}\ourmethodCC     & \cellcolor{Gray}\ding{51} & \cellcolor{Gray}$\textbf{69.82}_{0.32}$ & \cellcolor{Gray}$\textbf{57.20}_{0.45}$ & \cellcolor{Gray}$\textbf{71.97}_{0.46}$\\      \midrule
    \multirow{5}{*}{\rotatebox{90}{RESISC45}} &Zero-shot CLIP &  & \multicolumn{2}{c}{$62.67_{0.00}$} & $62.13_{0.00}$  \\ 
    &FPL                             & \ding{55} & $79.01_{0.55}$ & $70.85_{0.66}$ & $77.69_{0.83}$  \\
    &\cellcolor{Gray}\ourmethodCC    & \cellcolor{Gray}\ding{55} & \cellcolor{Gray}$\textbf{80.38}_{0.37}$ & \cellcolor{Gray}$\textbf{76.01}_{0.19}$ & \cellcolor{Gray}$\textbf{79.97}_{0.77}$     \\      
    &GRIP                             & \ding{51} & $81.53_{0.00}$ & $76.86_{0.00}$ & $86.88_{0.00}$  \\
    &\cellcolor{Gray}\ourmethodCC    & \cellcolor{Gray}\ding{51} & \cellcolor{Gray}$\textbf{87.75}_{0.29}$ & \cellcolor{Gray}$\textbf{80.88}_{0.86}$ & \cellcolor{Gray}$\textbf{89.73}_{1.73}$\\      \midrule
    \multirow{5}{*}{\rotatebox{90}{Flowers102}} &Zero-shot CLIP   &  & \multicolumn{2}{c}{$73.98_{0.00}$} & $73.05_{0.00}$  \\ 
    &FPL                             & \ding{55} & $\textbf{89.07}_{0.94}$ & $77.81_{0.30}$ & $91.84_{0.73}$  \\
    &\cellcolor{Gray}\ourmethodCC    & \cellcolor{Gray}\ding{55} & \cellcolor{Gray}$88.37_{0.39}$ & \cellcolor{Gray}$\textbf{82.98}_{0.14}$ & \cellcolor{Gray}$\textbf{96.65}_{0.08}$\\
    &GRIP                             & \ding{51} & $94.21_{0.00}$ & $82.33_{0.00}$ & $96.18_{0.00}$  \\
    &\cellcolor{Gray}\ourmethodCC    & \cellcolor{Gray}\ding{51} & \cellcolor{Gray}$\textbf{96.80}_{0.63}$ & \cellcolor{Gray}$\textbf{83.94}_{0.69}$ & \cellcolor{Gray}$\textbf{97.34}_{0.74}$\\
     \bottomrule[0.9pt]
    \end{tabular}}
\label{tab:diff_backbone}
\end{table}

\section{Limitations}

It is important to note that the performance of our method is dependent on the quality of the generated candidate pseudolabels. As such, it inherits the inherent limitation of pseudolabeling methods - the true label may not be included in the generated candidate pseudolabels. In future work, we aim to enhance the quality of candidate pseudolabels by refining the generation strategy and devising a method to better handle this situation.

\section{Conclusion}

In this paper, we have proposed a novel candidate pseudolabel learning (CPL) method to fine-tune vision-language models (VLMs), with abundant unlabeled data.
The key to our method lies in the strategy of generating suitable candidate pseudolabels, which contains both intra- and inter-instance label selection.
In this way, our generated candidate pseudolabels can offer two key advantages over conventional hard pseudolabels: enhanced accuracy in true label estimation and balanced representation across classes. 
Our extensive experiments, conducted across nine benchmark datasets and three learning paradigms validated the effectiveness of our method. This is particularly evident when the zero-shot capabilities of VLMs are not reliably applicable to downstream tasks.

\newpage
\section*{Acknowledgements}
Lei Feng is supported by the National Natural Science Foundation of China (Grant No. 62106028) and the Chongqing Overseas Chinese Entrepreneurship and Innovation Support Program. Feng Liu is supported by the Australian Research Council with grant numbers DP230101540 and DE240101089, and the NSF\&CSIRO Responsible AI program with grant number 2303037.

\section*{Impact Statement}
This paper presents work whose goal is to advance the field of Machine Learning. There are many potential societal consequences of our work, none of which we feel must be specifically highlighted here. 
\bibliography{0_main}
\bibliographystyle{icml2024}


\onecolumn
\newpage
\appendix

\centerline{\textbf{\Large Appendix for}}
\vspace{2mm}
\centerline{\textbf{\Large \emph{``Candidate Pseudolabel Learning: Enhancing Vision-Language Models by}}}
\vspace{2mm}
\centerline{\textbf{\Large \emph{Prompt Tuning with Unlabeled Data"}}}

In the appendix of this paper, we provide further details:
\begin{itemize}
    \item Elaboration on the iterative process for updating candidate pseudolabels (Appendix \ref{sec:appendix_a}).
    \item Exploration of an alternative approach for formulating a soft target from the candidate label set (Appendix \ref{sec:appendix_b}).
    \item Additional information about the datasets and hyperparameters used in our method (Appendix \ref{sec:appendix_c}).
    \item Presentation of experimental results derived from varying the primary hyperparameters and examination of the influence of the other hyperparameters (Appendix \ref{sec:appendix_D}).
\end{itemize}

\section{Details about Training Iterations}\label{sec:appendix_a}
In this section, we provide a detailed description of our iterative process for updating candidate pseudolabels, as outlined in \cref{alg:curriculum_labeling}. Specifically, in each iteration, after generating candidate pseudolabels for each instance, we filter out instances where the candidate set is empty, forming $D_{\rm temp}$. Subsequently, for each class, we select the $K_t$ instances with the highest confidence scores, add these instances to the training set $D_{\rm T}$, and simultaneously remove these instances from $D_{\rm UL}$ to avoid repeated selection. Through this step, we ensure that the number of instances for each class does not exceed $K_t$ and output the training set $D_{\rm T}$ for the current iteration.

Simultaneously, to utilize more training data, after the iteration is completed, we increase $K_t$ by $\Delta$ so that more unlabeled data can be utilized in the next iteration. This iteration process is then repeated until we reach the maximum number of iterations $T$. Normally, The increment for each category’s quantity per iteration is set to $\Delta=\frac{|D_{\rm UL}|}{T}$. 
This process is akin to curriculum learning, as it starts training with simple and reliable instances and gradually increases the difficulty. The emphasis is on updating the candidate pseudolabels corresponding to unlabeled data before each iteration and reinitializing the learnable parameters of the model. 

The overall iteration process in the CPL and prior work \cite{menghini2023enhancing} both adhere to the above-mentioned iterative update strategy, ensuring fair comparison. At the same time, we have made some simple modifications to make it more suitable for our candidate pseudolabels.

Specifically, to select a designated number of instances for class $c$ (class-wise Top-$K_t$ selection), the approach adopted in previous work~\cite{menghini2023enhancing} utilized the confidence scores of all potential labels for the ranking process. In contrast, we confine our ranking and selection to only the labels included in candidate label set $S$. In other words, we only involve all confidence scores of the instances for which $c \in S$ in the ranking process. This makes the improved top-$K$ selection method more appropriate for candidate pseudolabels, as we should prioritize the categories in the candidate label set during the selection process and exclude the influence of non-candidate categories.

\begin{algorithm}[h]
    \footnotesize
    \renewcommand{\algorithmicrequire}{\textbf{Input:}}
    \renewcommand{\algorithmicensure}{\textbf{Output:}}
    \caption{Top-$K$ Selection Process in Each Iteration}
    \label{alg:curriculum_labeling}
    \begin{algorithmic}[1]
        \REQUIRE Total iteration number $T$, an unlabeled set $D_{\rm UL} := \{\bs{x}_i\}_{i=1}^N$, the number of instances $K_{t}$ that should be selected in iteration $t$ for all classes. 
        The increment number between two iterations is $\Delta$. \\
        \ENSURE The training set $D_{\rm T}$ with candidate psuedolabels 
        \FOR{$t \in 1,...,T$}
        \STATE Initialize the training set $D_{\rm T} := \varnothing$
            \STATE Generate candidate pseudolabels $S$ for each unlabeled instance $\boldsymbol{x}_{i}$ according to two label selection strategies (with Eq. (\ref{eq:eq_S_intra}) (\ref{eq:eq_S_inter})).\\
            \STATE Refine the set of instance-candidate pairs $D_{\rm temp} := \{(\bs{x}_i, S_i)\}_{i=1}^M$ by filtering the sample where $S_i=\varnothing$ \\
                \FOR{$c \in [C]$} 
                    \STATE $Q := |D_{\rm temp}|$ \hfill {\color{gray} \small  $\triangleright$ $Q$ is the number of instances in current $D_{\rm temp}$}
                    \STATE $\mathcal{V}_c := \{p_{ic}| c \in S_i\}_{i=1}^Q$ \hfill {\color{gray} \small  $\triangleright$ Collect the corresponding confidence scores when $c$ is contained in the set $S_i$}
                    \STATE $D_{\rm T} \leftarrow D_{\rm T} \bigcup \{(\bs{x}_i, S_i) | p_{ic} \in \text{Top-}K_{t}(\mathcal{V}_c)\}_{i=1}^Q$         \hfill {\color{gray} \small  $\triangleright$ Select top-$K_{t}$ instances according to the candidate labels' confidence scores}
                    \IF{sample $(\bs{x}_i, S_i)$ is selected}
                        \STATE $D_{\rm temp} \gets D_{\rm temp} \backslash (\bs{x}_i, S_i)$ \\
                    \ENDIF
                \ENDFOR
            \STATE $K_{t+1} := K_{t} + \Delta$ 
            \STATE Transform each candidate set $S$ in $D_{\rm T}$ into training target $\bs{s}$.  \\
            \STATE Return the training set $D_{\rm T}$.
        \ENDFOR
    \end{algorithmic}
\end{algorithm}

\begin{table}[h]   
    \centering
    \small
    \caption{Comparison results of top-1 test accuracy (\%) on UL task with textual prompts tuning. The main difference here is using a soft target or not.}
    \vspace{-2mm}

    \label{tab:compare_soft_target}
     \resizebox{0.40\linewidth}{!}{
    \begin{tabular}{l|cccc}
    \toprule[0.9pt]
    \multicolumn{1}{c|}{Methods}   & EuroSAT & DTD  & Flowers102 \\\midrule
    \ourmethodCC$_{\mathrm{CC}}$             & $67.26$ & $51.91$ & $72.90$  \\
    \ourmethodCC$_{\mathrm{RC}}$             & $66.38$ & $51.58$ & $71.91$  \\
    \ourmethodCC$_{\mathrm{CAV}}$            & $66.91$ & $49.31$ & $72.06$  \\
    \ourmethodCC$_{\mathrm{LW}}$             & $67.15$ & $52.23$ & $72.33$  \\ \midrule
    \ourmethodCC$_{\mathrm{Soft\,\,CE}}$     & $56.85$ & $49.53$ & $71.87$  \\
    \ourmethodCC$_{\mathrm{Soft\,\,CC}}$             & $65.58$ & $50.98$ & $72.58$ \\
    \ourmethodCC$_{\mathrm{Soft\,\,RC}}$             & $65.89$ & $50.83$ & $71.89$ \\
    \ourmethodCC$_{\mathrm{Soft\,\,CAV}}$            & $64.90$ & $48.78$ & $71.72$ \\
    \ourmethodCC$_{\mathrm{Soft\,\,LW}}$             & $65.74$ & $51.04$ & $72.15$ \\
     \bottomrule[0.9pt]
    \end{tabular}}
    \vspace{-1mm}
\end{table}

\section{Discussion on Soft Pseudolabels}\label{sec:appendix_b}   
In this section, we explore an alternative approach to utilizing the candidate label set $S$ as the learning target, specifically by formulating a soft target derived from $S$. This approach aims to provide a more comprehensive evaluation of our method.

Unlike candidate pseudolabels, which treat all candidate labels equally, soft pseudolabels utilize normalized confidence scores from the previous model's prediction as the soft target for the subsequent iteration. Specifically, the confidence scores from the preceding model's output for each category $c \in S$ are used as the criteria for defining the soft target. 

Let $\boldsymbol{\hat{y}} = (y_{1}, y_{2}, y_{3}, \cdots, y_{C})$ represents the soft target for unlabeled instance $\boldsymbol{x}$. For a category $c \in [C]$, the corresponding value $y_c$ in soft target can be calculated as follows: 
\begin{equation}
    \label{Eq:soft_target}
    y_c = 
    \begin{cases}
      \frac{P\left(y=c | \boldsymbol{x}\right)}{\sum_{k \in S} P\left(y=k | \boldsymbol{x}\right)} = \frac{p_{c}}{\sum_{k \in S} p_{k}}, & \text{if}\ c \in S \\
      0, & \text{otherwise}
    \end{cases},
\end{equation}
where $S$ is the candidate pseudolabel set for instance $\boldsymbol{x}$. We compared the performance between this soft target generation method and CPL which treats all candidate labels equally, as shown in \cref{tab:compare_soft_target}. 

In \cref{tab:compare_soft_target}, ``Soft CE" refers to the approach where the soft target is obtained using \cref{Eq:soft_target} before each iteration, and training is conducted using the cross-entropy loss. Similarly, ``Soft CC" refers to the method where the soft target is obtained in the same manner, but training is executed using the CC objective function. All other experimental settings are the same as those in \cref{sec4.1:exp_setting}.

From the table, we observe that methods not utilizing the soft target exhibit marginally superior performance on the three datasets. This suggests that directly using candidate pseudolabels as the training target can more effectively enhance the performance of CLIP when fine-tuning on downstream tasks, especially when the zero-shot capabilities of CLIP are relatively weak on these datasets.
We posit that this may be attributed to the fact that the utilization of a soft target, which contains more information, might inadvertently reinforce the model's previous erroneous predictions and class bias, thus exacerbating the impact of confirmation bias and hindering subsequent learning process. Therefore, in the main text, we implement our CPL by treating all candidate labels equally.

\section{Experimental Details}\label{sec:appendix_c}

\subsection{Comparison Methods}
In this section, we provide a detailed introduction to the methods included in our experiments, which are divided into two categories. The first category encompasses strategies for fine-tuning CLIP under limited data conditions, including:

\begin{itemize}[leftmargin=6mm]
\setlength\itemsep{0mm}
\vspace{-3mm}
    \item \textbf{Few-pseudolabels (FPL)} \cite{menghini2023enhancing}: This approach generates offline pseudolabels by selecting the top-$K$ samples with the highest confidence for each class from CLIP zero-shot predictions, performed only once.
    \item \textbf{Grow and Refine Iteratively Pseudolabels (GRIP)} \cite{menghini2023enhancing}: GRIP maintains class balance by selecting the top-$K$ samples at each iteration for each class, with $K$ progressively increasing after each iteration. The key distinction between GRIP and FPL lies in GRIP's iterative pseudolabel updates and the incremental increase of samples for each class at each iteration.  
    \item \textbf{Unsupervised Prompt Learning (UPL)} \cite{huang2022unsupervised}: Align with FPL, UPL employs the most confident samples for each class and generates offline pseudolabels to learn text prompts through the CLIP text encoder.
    \item \textbf{CLIP-PR} \cite{kahana2022improving}: This method optimizes an adapter atop the CLIP vision encoder. It uses label distribution priors from the training set of downstream datasets and generates offline pseudolabels only once. 
    \item \textbf{LaFTer} \cite{mirza2023lafter}: This method utilizes an unlabeled image collection and a set of text descriptions generated by a Large Language Model (LLM) to fine-tune CLIP with online pseudolabels. Notably, it also generates hard pseudolabels and employs a consistency regularization strategy \cite{sohn2020fixmatch} to learn from unlabeled data. 
\end{itemize}

The second category pertains to the loss functions in partial-label learning:
\begin{itemize}[leftmargin=6mm]
\setlength\itemsep{0mm}
\vspace{-3mm}
    \item \textbf{Classifier-Consistent (CC) \& Risk-Consistent (RC)} \cite{feng2020provably}: These two methods are designed for partial-label learning. They develop two novel methods that are guaranteed to be provably consistent when dealing with learning from a candidate set of labels.
    \item \textbf{Class Activation Value (CAV)} \cite{zhang2021exploiting}
: This method introduces the class activation value as a versatile tool to select the true label. It identifies the class with the maximum CAV for model training.
    \item \textbf{Leveraged Weighted (LW) Loss} \cite{wen2021leveraged}: The leveraged weighted loss function introduces a leverage parameter to balance the losses on partial labels and non-partial ones. 
\end{itemize}

\subsection{Task Introduction}\label{sec:appendix_c1}
We provide a detailed explanation of experimental settings for three learning paradigms, in line with \cite{menghini2023enhancing}.
\begin{itemize}[leftmargin=6mm]
\setlength\itemsep{0mm}
\vspace{-3mm}
    \item For \textbf{Semi-Supervised Learning (SSL)}, access to labeled data is limited. We assess the impact of pseudolabels in scenarios with a few labeled data and abundant unlabeled data, using two labeled samples per class.
    \item For \textbf{Unsupervised Learning (UL)}, we only have access to unlabeled data. In this scenario, we initially rely on the zero-shot predictions of CLIP to obtain all pseudolabels without any manual annotation.
    \item For \textbf{Transductive Zero-Shot Learning (TRZSL)}, labeled data for certain target classes (seen classes) are provided in the downstream dataset. We set the ratio of seen to unseen classes at 62-38, with all pseudolabels generated from unseen classes. It is noteworthy that in TRZSL, we report the harmonic mean of the accuracies of seen and unseen classes.
\end{itemize}

\begin{table*}[h]
\centering
\caption{Detailed settings for experiments in Section \ref{sec:Experiments}.} 
\vspace{-4mm}
\resizebox{0.9\textwidth}{!}{
\setlength{\tabcolsep}{5mm}{
\begin{tabular}{l|c|c|c|c|c|c}
\toprule[1.1pt]
      & Flowers102   & RESISC45     & DTD  & CUB  & EuroSAT  & FGVCAircraft \\  \midrule
\multicolumn{5}{l}{\textbf{Statistic data}} \\ \midrule
\quad Class number    & 102           & 45           & 47     & 200  & 10  & 100\\ 
\quad Training set size   & 2040       & 6300       & 3760 & 5594  & 27000  & 6667\\ 
\quad Testing set size    & 6149    & 25200     & 1880  & 5794  & 5000  & 3333    \\ \midrule
\multicolumn{5}{l}{\textbf{Training procedure}} \\ \midrule
\quad Network         & \multicolumn{6}{c}{ViT-B / 32} \\ \cline{2-7}
\quad Batch size      & \multicolumn{6}{c}{64} \\ \cline{2-7}
\quad Epoch           & \multicolumn{6}{c}{50 where first two epochs are set for warmup} \\ \cline{2-7}
\quad Optimizer       & \multicolumn{6}{c}{SGD} \\ \cline{2-7}
\quad Momentum        & \multicolumn{6}{c}{0.9} \\ \cline{2-7}
\quad Learning rate (LR)  & \multicolumn{6}{c}{0.02} \\ \cline{2-7}
\quad Weight decay    & \multicolumn{6}{c}{5e-2} \\ \cline{2-7}
\quad LR scheduler    & \multicolumn{6}{c}{CosineAnnealingLR} \\\midrule
\multicolumn{5}{l}{\textbf{Hyperparameters}}  \\ \midrule
\quad  $\alpha$ in intra-instance label selection & 0.60 & 0.90 & 0.75 & 0.75 & 0.75 & 0.90 \\ \cline{2-7}
\quad  $\beta$ in inter-instance label selection & 0.99 & 0.97 & 0.95 & 0.99 & 0.80 & 0.97 \\ 
\bottomrule[1.1pt]
\end{tabular}
}}
\label{tab:train_setup}
\vspace{-4mm}
\end{table*}

\subsection{Datasets and Hyperparameters}\label{sec:appendix_c2}
In this section, we provide additional visualization and details regarding the datasets and hyperparameters used in CPL.

\textbf{Setup}.
We provide the statistical data for six datasets and the complete experimental setup in Table \ref{tab:train_setup}. 

\begin{figure}[h]
    \centering 
    \includegraphics[width=0.85\linewidth]{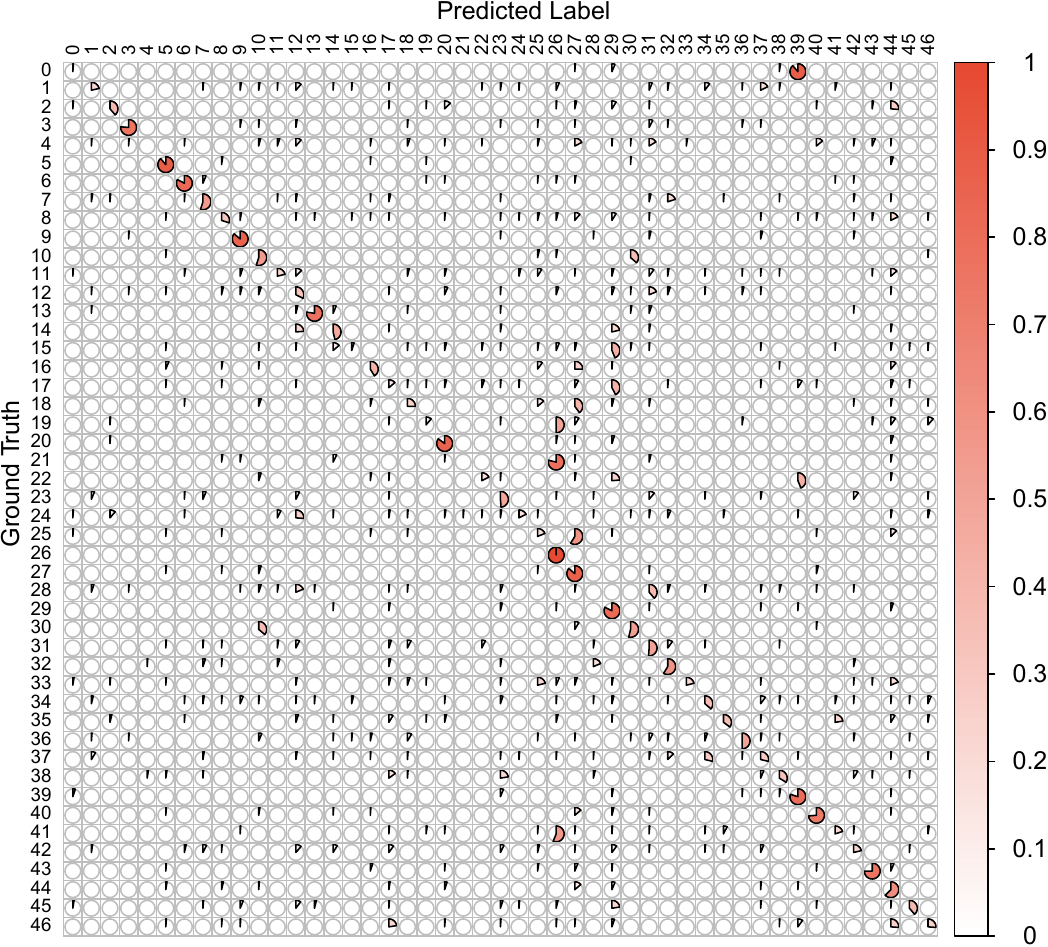}
        \caption{Confusion matrix illustrating the correlation between true labels and predicted hard pseudolabels on DTD. }
        \vspace{-4mm}
    \label{fig:matrix_dtd}
\end{figure}

\textbf{Additional Dataset Visualizations}. 
In addition to the pilot experiment in Figure \ref{fig:matrixa} in the main text, which reveals the low label estimation accuracy and class bias issues of hard pseudolabels on EuroSAT, we also visualize the confusion matrix on the DTD dataset in a similar manner, as shown in Figure \ref{fig:matrix_dtd}. This visualization reveals similar issues on the DTD dataset.

\section{Additional Experimental Results}\label{sec:appendix_D}
In this section, we present the experimental results derived from our method when varying the two primary hyperparameters ($\alpha$, $\beta$) and hyperparameter $\lambda$ which controls the balance between labeled and unlabeled data for SSL and TRZSL tasks.

\textbf{Ablation Studies for the Trade-off Coefficient $\lambda$}. For the SSL and TRZSL tasks, we conduct ablation studies on the hyperparameter $\lambda$, as shown in Table \ref{tab:compare_lamda_ssl} and Table \ref{tab:compare_lamda_trzsl}. We observe that, except for a few datasets (e.g., Flowers102), most datasets do not exhibit excessive dependency on the hyperparameter $\lambda$. Therefore, for SSL and TRZSL, we consistently set $\lambda$ to 1 in the main text to avoid the impact of over-parameterization.

\begin{table}[h]   
    \centering
    \small
    \caption{Comparison of top-1 test accuracy (\%) on \textbf{SSL} tasks with textual prompt tuning, illustrating the influence of $\lambda$. Here, we use all the unlabeled data for fine-tuning CLIP.
    }
    \label{tab:compare_lamda_ssl}
     \resizebox{0.75\linewidth}{!}{
    \begin{tabular}{l|cccccc}
    \toprule[0.9pt]
    \multicolumn{1}{c|}{Methods}  & Flowers102 & RESISC45  & DTD & CUB & EuroSAT  & FGVCAircraft \\\midrule
    Zero-shot CLIP                & $63.67$ & $54.48$ & $43.24$  & $51.82$ & $32.80$  & $17.58$ \\ \midrule  
    \ourmethodCC$_{\lambda=0.50}$           & $90.09$ & $80.97$ & $58.96$ & $58.11$ & $76.71$  & $22.76$ \\
    \ourmethodCC$_{\lambda=0.75}$           & $90.28$ & $81.76$ & $60.13$ & $58.46$ & $76.96$  & $22.07$ \\
    \ourmethodCC$_{\lambda=1.00}$           & $89.66$ & $80.98$ & $61.21$ & $58.53$ & $77.51$  & $22.48$ \\
    \ourmethodCC$_{\lambda=1.25}$           & $89.68$ & $81.37$ & $59.31$ & $58.42$ & $77.60$  & $22.86$ \\
    \ourmethodCC$_{\lambda=1.50}$           & $88.77$ & $81.64$ & $61.28$ & $58.28$ & $77.33$  & $22.65$ \\
     \bottomrule[0.9pt]
    \end{tabular}}
    \vspace{3mm}
\end{table}

\begin{table}[h]   
    \centering
    \small
    \caption{Comparison of top-1 test accuracy (\%) on \textbf{TRZSL} tasks with textual prompt tuning, illustrating the influence of $\lambda$. Here, we use all the unlabeled data for fine-tuning CLIP.
    }
    \label{tab:compare_lamda_trzsl}
     \resizebox{0.75\linewidth}{!}{
    \begin{tabular}{l|cccccc}
    \toprule[0.9pt]
    \multicolumn{1}{c|}{Methods}  & Flowers102 & RESISC45  & DTD & CUB & EuroSAT  & FGVCAircraft \\\midrule
    Zero-shot CLIP                & $63.40$ & $54.46$ & $43.45$  & $51.57$ & $30.54$  & $17.86$ \\ \midrule
    \ourmethodCC$_{\lambda=0.50}$           & $90.01$ & $85.95$ & $67.31$ & $65.28$ & $93.37$  & $32.60$ \\
    \ourmethodCC$_{\lambda=0.75}$           & $86.91$ & $84.91$ & $67.77$ & $65.25$ & $93.11$  & $31.47$ \\
    \ourmethodCC$_{\lambda=1.00}$           & $87.35$ & $85.85$ & $68.00$ & $63.94$ & $93.78$  & $30.26$ \\
    \ourmethodCC$_{\lambda=1.25}$           & $88.98$ & $86.10$ & $67.82$ & $64.84$ & $94.01$  & $30.42$ \\
    \ourmethodCC$_{\lambda=1.50}$           & $87.13$ & $85.96$ & $68.11$ & $64.81$ & $93.56$  & $30.05$ \\
     \bottomrule[0.9pt]
    \end{tabular}}
    \vspace{3mm}
\end{table}

\begin{figure}[H]
\centering     
\subfigure[SSL on DTD]{\label{fig:ssl_1}\includegraphics[width=84mm]{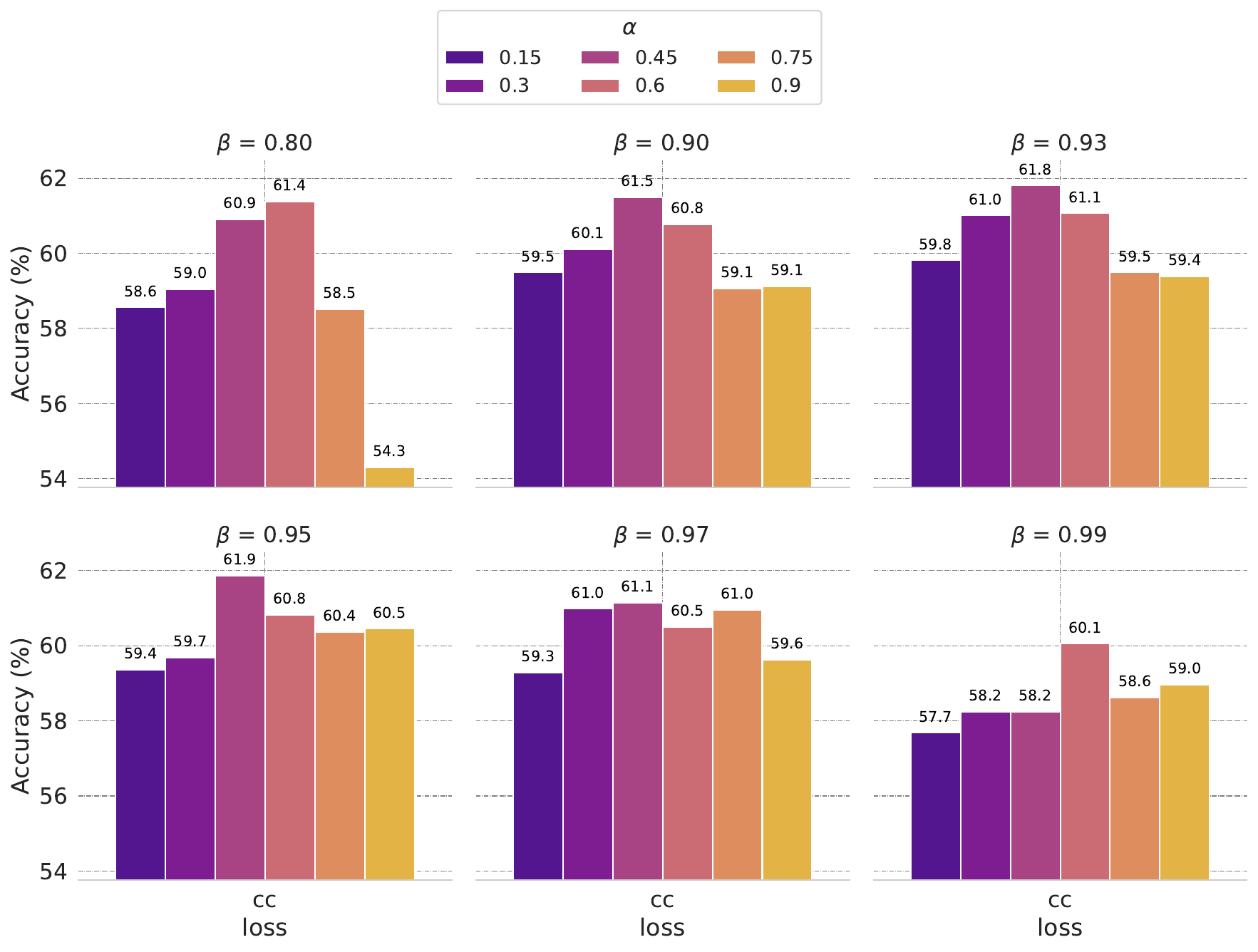}}
\hspace{1mm}
\subfigure[SSL on DTD]{\label{fig:ssl_2}\includegraphics[width=84mm]{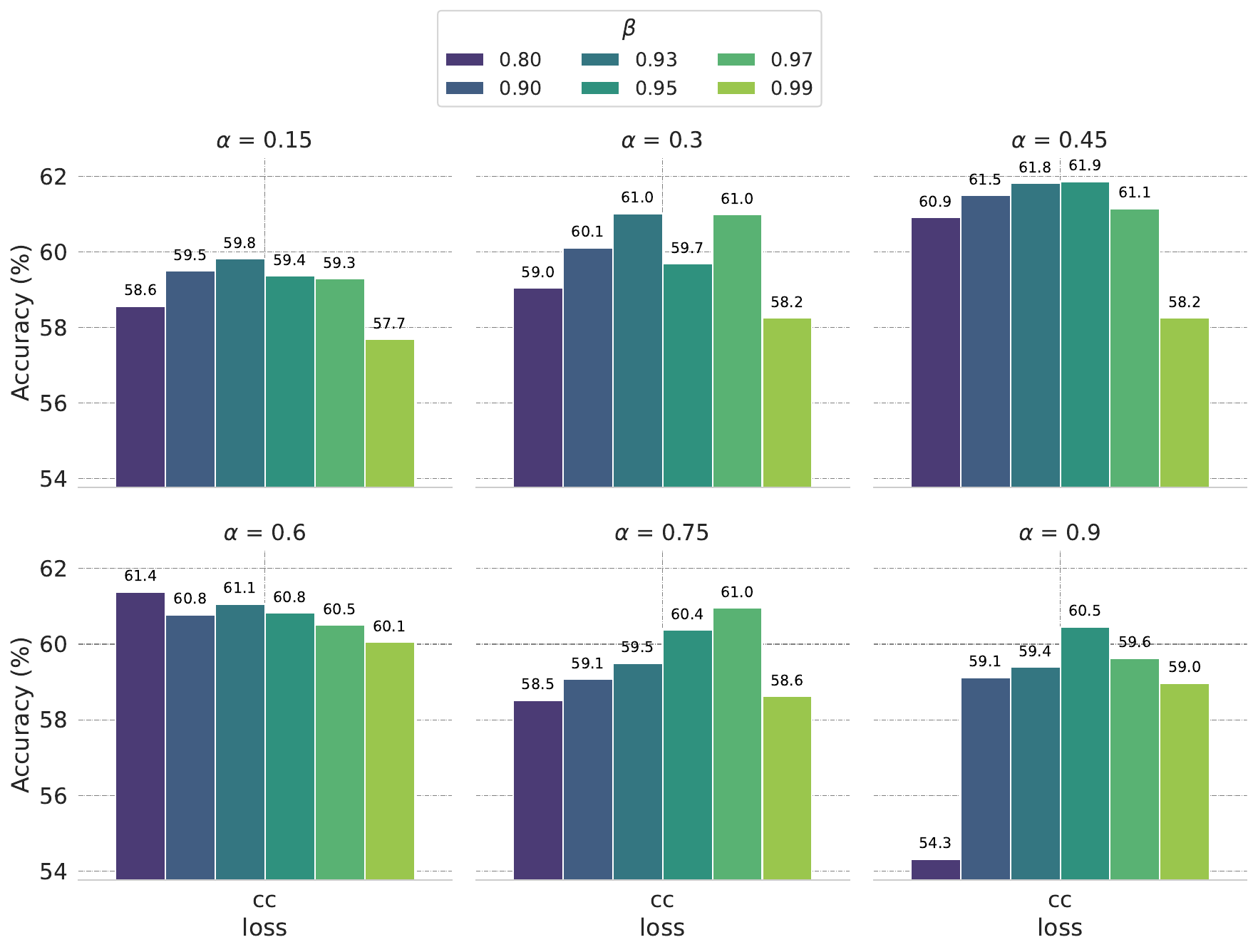}}\\
\vspace{-3mm}
\caption{(a) Illustrates the impact of the parameter $\alpha$ under various settings of $\beta$. 
(b) Illustrates the impact of the parameter $\beta$ under various settings of $\alpha$. 
}
\vspace{-2mm}
\label{fig:hyperparams_subs_dtd_ssl}
\end{figure}

\begin{figure}[t]
\centering     
\subfigure[UL on DTD]{\label{fig:ul_1}\includegraphics[width=84mm]{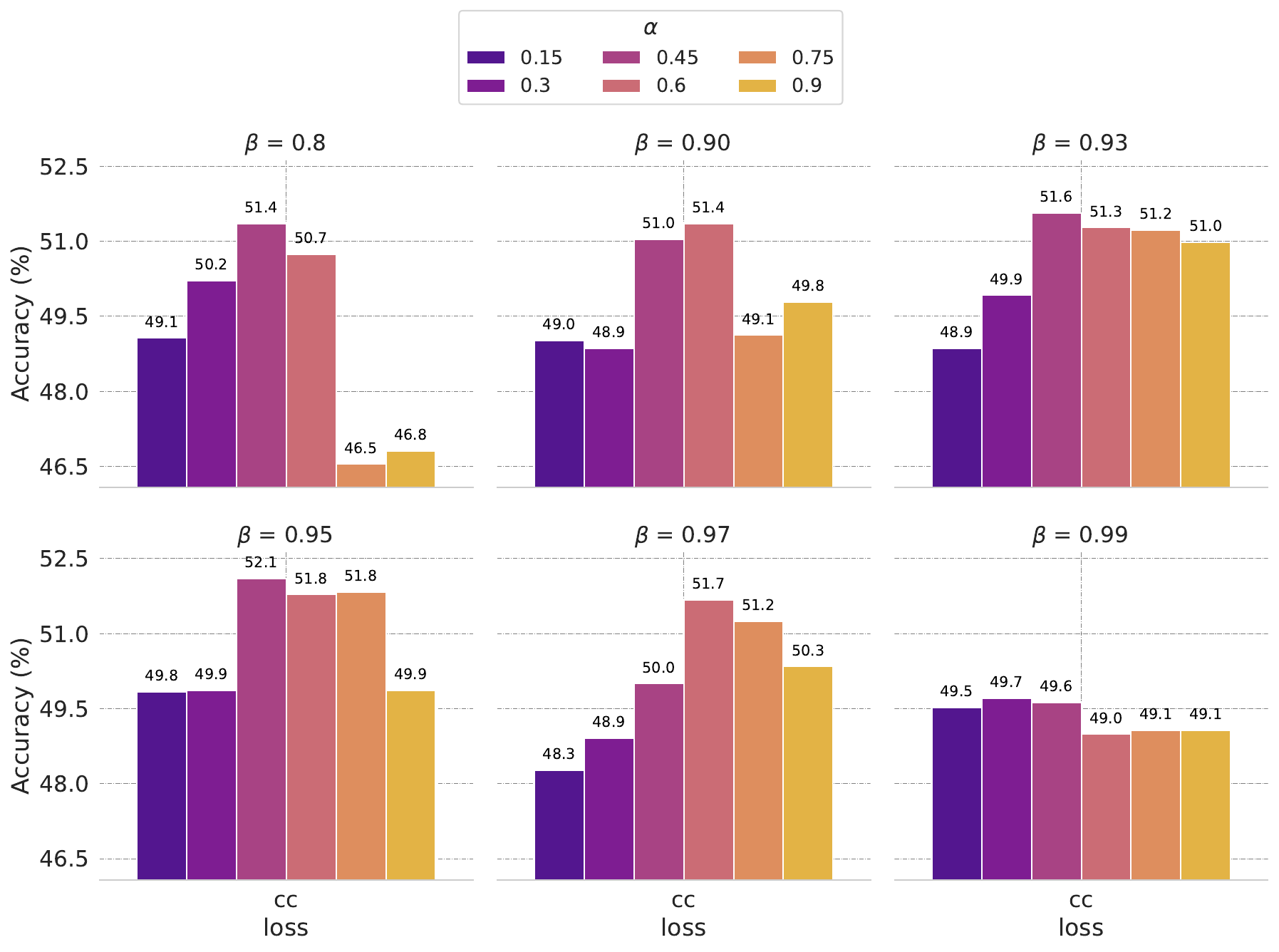}}
\hspace{1mm}
\subfigure[UL on DTD]{\label{fig:ul_2}\includegraphics[width=84mm]{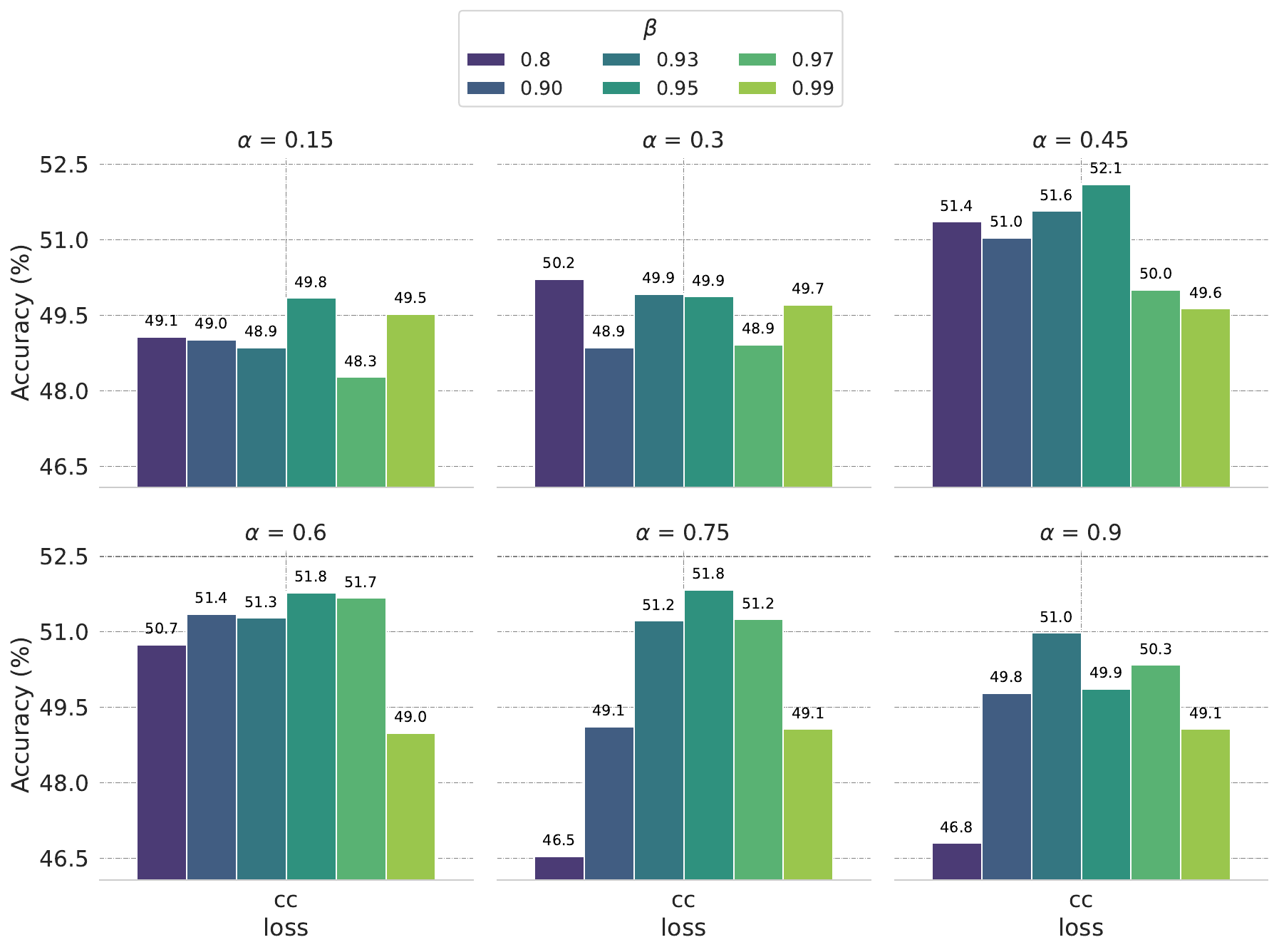}}\\
\vspace{-3mm}
\caption{(a) Illustrates the impact of the parameter $\alpha$ under various settings of $\beta$. 
(b) Illustrates the impact of the parameter $\beta$ under various settings of $\alpha$. 
}
\vspace{-2mm}
\label{fig:hyperparams_subs_dtd_ul}
\end{figure}

\begin{figure}[t]
\centering     
\subfigure[TRZSL on DTD]{\label{fig:trzsl_1}\includegraphics[width=84mm]{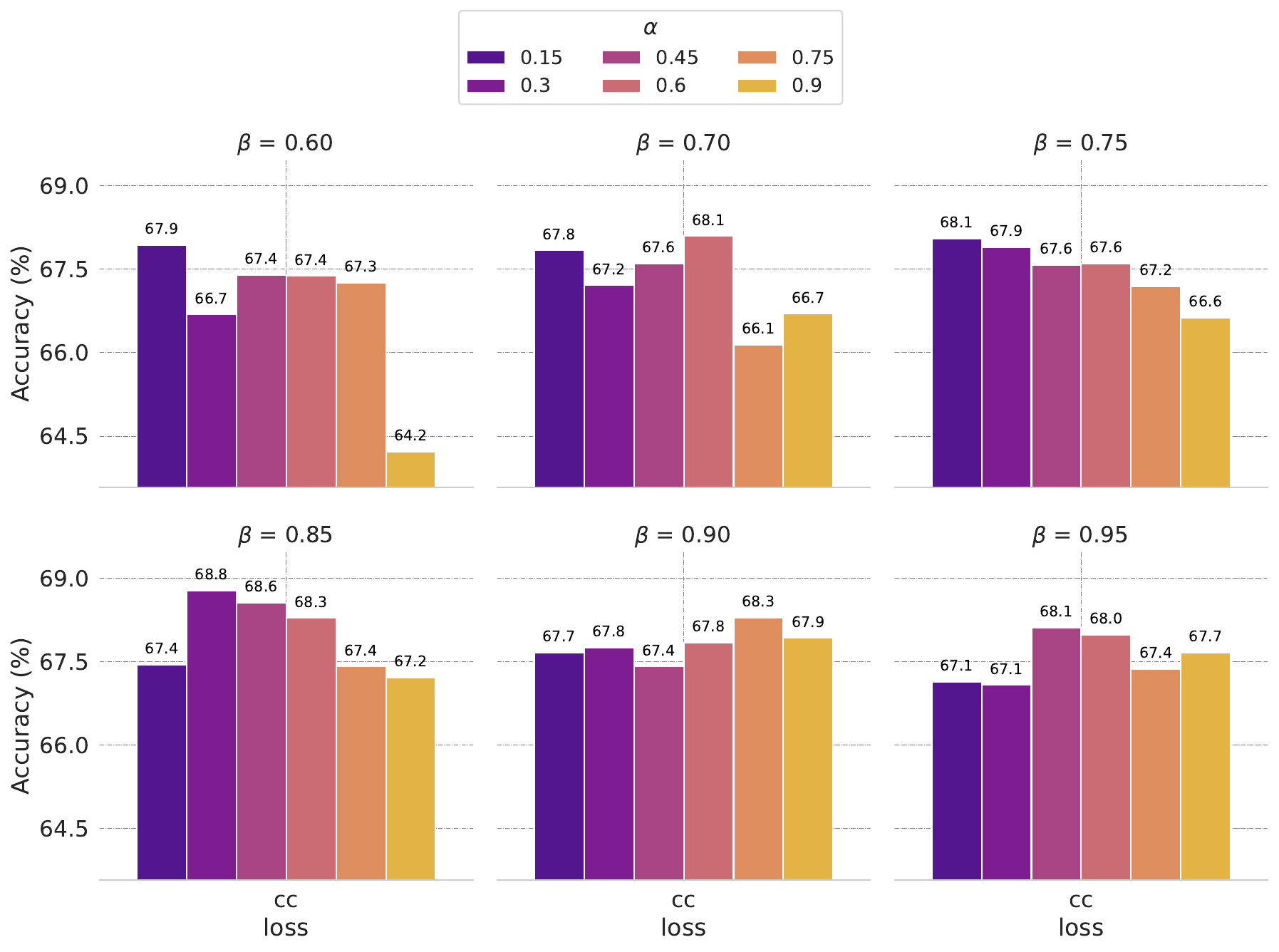}}
\hspace{1mm}
\subfigure[TRZSL on DTD]{\label{fig:trzsl_2}\includegraphics[width=84mm]{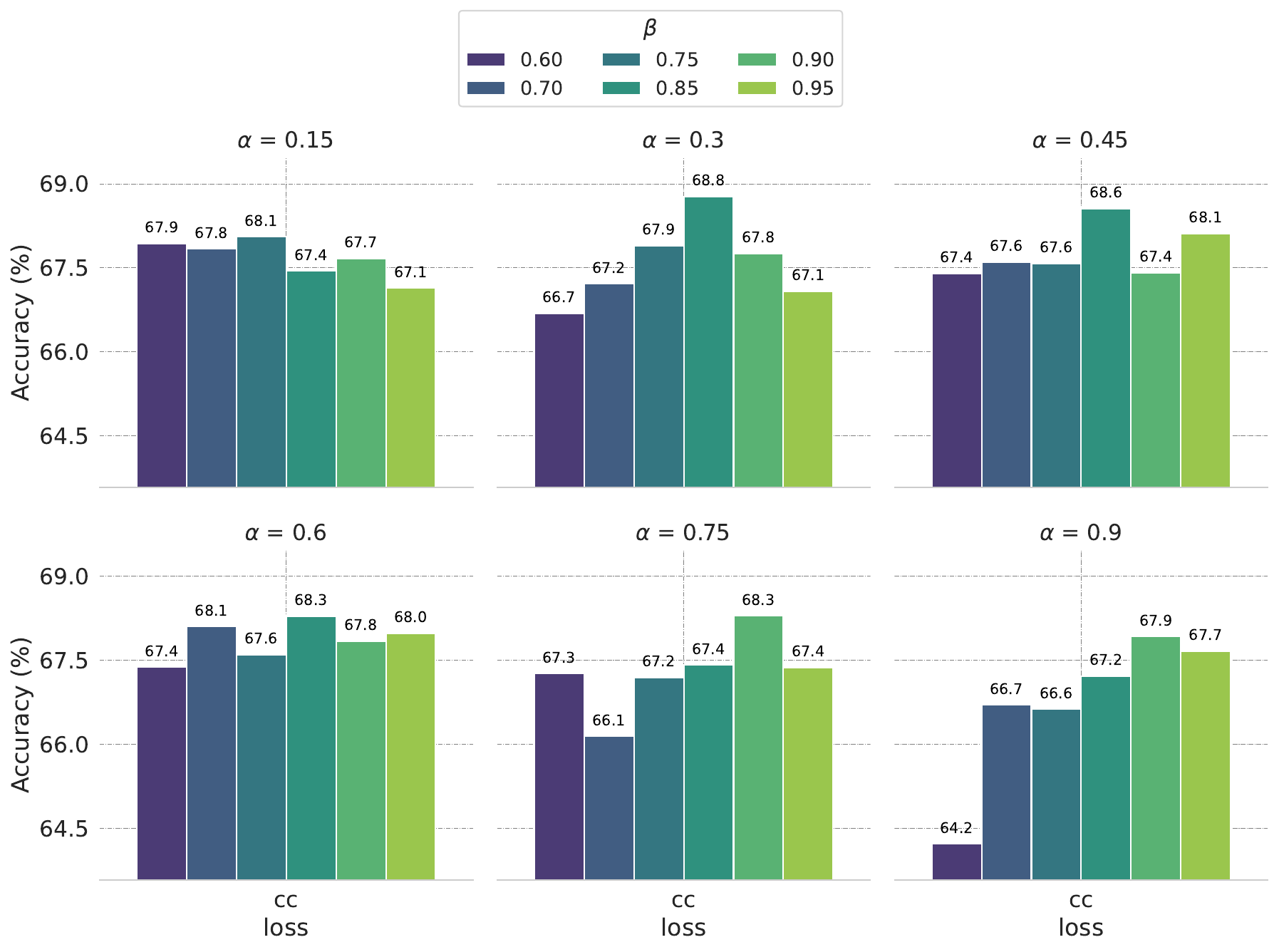}}\\
\vspace{-3mm}
\caption{(a) Illustrates the impact of the parameter $\alpha$ under various settings of $\beta$. 
(b) Illustrates the impact of the parameter $\beta$ under various settings of $\alpha$. 
} 
\label{fig:hyperparams_subs_dtd_trzsl}
\vspace{-2mm}
\end{figure}

\textbf{Grid Search for Hyperparameter Selection}. We conduct ablation studies to examine the influence of hyperparameters more comprehensively. We set $\alpha \in \{0.15, 0.30, 0.45, 0.60, 0.75, 0.90\}$ and $\beta \in \{0.80, 0.90, 0.93, 0.95,0.97, 0.99\}$ for SSL and UL, and $\beta \in \{0.60, 0.70, 0.75, 0.85,0.90, 0.95\}$ for TRZSL.The results of these experiments for the three tasks on the DTD dataset are depicted in Figures \ref{fig:hyperparams_subs_dtd_ssl}, \ref{fig:hyperparams_subs_dtd_ul}, and \ref{fig:hyperparams_subs_dtd_trzsl}. The results indicate that our method is robust to changes in these hyperparameters in a range (especially in TRZSL and SSL), and can achieve competitive performance across a wide range of settings. 

\end{document}